\documentclass[oneside,english,reqno]{amsart}

\usepackage[LGR,T1]{fontenc}
\usepackage[utf8]{inputenc}
\usepackage{color}
\usepackage{array}
\usepackage{multirow}
\usepackage{amsbsy}
\usepackage{amstext}
\usepackage{amsthm}
\usepackage{amssymb}
\usepackage{graphicx}
\usepackage{esint}

\makeatletter

\providecommand{\tabularnewline}{\\}

\numberwithin{equation}{section}
\numberwithin{figure}{section}

\usepackage[notquote]{hanging}

\makeatother

\usepackage{babel}

\begin{document}
\title{Flat-topped Probability Density Functions for Mixture Models}
\author{Osamu Fujita}
\address{Department of Educational Collaboration, Osaka Kyoiku University, 4\textendash 698\textendash 1
Asahigaoka, Kashiwara, Osaka 582\textendash 8582, Japan}
\email{fuji@cc.osaka-kyoiku.ac.jp}

% \begin{abstract}
\begin{abstract}
This paper investigates probability density functions (PDFs) that are continuous
everywhere, nearly uniform around the mode of distribution, and adaptable
to a variety of distribution shapes ranging from bell-shaped to rectangular.
From the viewpoint of computational tractability, the PDF based on
the Fermi-Dirac or logistic function is advantageous in estimating
its shape parameters. The most appropriate PDF for $n$-variate
distribution is of the form: $p\left(\mathbf{x}\right)\propto\left[\cosh\left(\left[\left(\mathbf{x}-\mathbf{m}\right)^{\mathsf{T}}\boldsymbol{\Sigma}^{-1}\left(\mathbf{x}-\mathbf{m}\right)\right]^{n/2}\right)+\cosh\left(r^{n}\right)\right]^{-1}$
where $\mathbf{x},\mathbf{m}\in\mathbb{R}^{n}$, $\boldsymbol{\Sigma}$
is an $n\times n$ positive definite matrix, and $r>0$ is a shape
parameter. The \textit{flat-topped} PDFs can be used as a component
of mixture models in machine learning to improve goodness of fit and
make a model as simple as possible.  
\end{abstract}

% \keywords
\keywords{flat-topped distribution, generalized Fermi-Dirac distribution, hyperbolic function, compound distribution, mixture model, generalized EM algorithm}
\maketitle

% section{Introduction}
\section{Introduction}

In machine learning, mixture models~\cite{Everitt,Bishop,McLachlan}
are valuable for modeling and analyzing complex real-world data. The
Gaussian mixture model (GMM) is widely used due to its simplicity
and fundamentality. It is easy to estimate the parameters of a single
Gaussian (normal) distribution using maximum likelihood (ML) estimation,
which simplifies the M-step of the expectation\textendash maximization
(EM) algorithm~\cite{Dempster}. Besides, exponential families have
conjugate priors that help derive analytical expressions in maximum
a posteriori (MAP) estimation and Bayesian inference such as Variational
Bayes (VB)~\cite{Attias}. 

In practice, however, bell-shaped distributions are not always appropriate
for modeling real data sets. There can be a variety of data coming
from non-Gaussian distributions. A uniform (rectangular) distribution
$\mathcal{U}\left(a,b\right)$ will be proper for data points uniformly
distributed in line, area, or volume elements, which often appear
in spatial analysis. It attaches importance to the distribution boundaries,
$a$ and $b$, contrasted with the normal distribution characterized
by the central tendency and deviation from the mean. Both concepts
are essential to develop various methods for data clustering and classification.
The uniform distribution is also fundamental but has drawbacks for
the components of mixture models. Its probability density function
(PDF) is zero outside the interval $[a,b]$ and discontinuous
at $a$ and $b$, which are disadvantageous for ML estimation and
numerical optimization algorithms. To avoid these inconveniences,
it should be modified to have smooth steps at its boundaries.

Some univariate PDF families have desirable properties. For example,
the generalized normal distribution~\cite{Varanasi} (or the exponential
power distribution~\cite{Crooks,Liu}), the generalized Cauchy distribution~\cite{Rider}
(or the generalized Pearson VII distribution~\cite{Crooks,Pearson}),
and the Ferreri distribution~\cite{Ferreri} are supported on the
whole real line, continuous, unimodal, and can be \textit{flat-topped}
about the mode. They are used not only for data clustering via mixture
models~\cite{Dang} but also for the studies of laser beam shapes~\cite{Shealy},
uncertainty in measurements~\cite{Blazquez}, and noise distributions~\cite{Tan}.
Unfortunately, their shape parameters are difficult to estimate. In
practice, alternative PDFs have been proposed for the mixtures of
rectangles~\cite{Pelleg,Alivanoglou}, though there remains a need
for more detailed research. 

In this paper, we study a variety of \textit{flat-topped} PDFs useful for finite mixture models. The condition of flatness is described in the following section. The \textit{flat-topped} PDFs based on
various combinations of sigmoid functions or generalization of the
Cauchy distribution are categorized into four general types in Section~3
and illustrated with some specific forms in Section~4. From the viewpoint
of computational tractability, the combination of the logistic functions
is most appropriate for building mixture models. Furthermore, a generalized
Fermi-Dirac distribution and its variant using hyperbolic functions
are advantageous for modeling multivariate elliptical distributions.
We also discuss the ML estimation of model parameters using an iterative
method in Section~5, a practical procedure using the generalized
EM algorithm~\cite{Dempster} to build mixture models in Section~6,
and the usefulness of the \textit{flat-topped} PDFs with some simulation
examples in Section~7.

% section{Preliminaries}
\section{Preliminaries}

To deal with the vague concept of ``\textit{flat-topped}'' PDF,
we determine quantitative criteria for describing its property. We
verify that the generalized normal distribution satisfies this property
under certain conditions and see how difficult it is to estimate
its shape parameter.

% subsection{Condition of flat-topped PDF}
\subsection{Condition of flat-topped PDF}

\begin{figure}[h]
\begin{centering}
\includegraphics[scale=0.9]{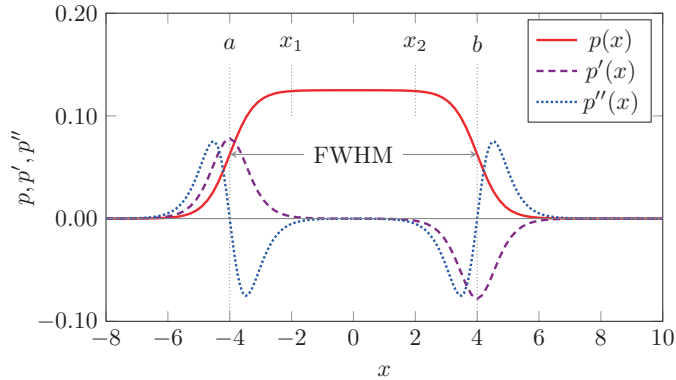}
\caption{The PDF of a typical \textit{flat-topped} distribution and its derivatives}
\label{fig:typical-flat-topped}
\end{centering}
\end{figure}

Let $p\left(x\right)$ be a PDF that is continuous for all $x\in\mathbb{R}$.
If it is twice differentiable, let $p^{\prime}\left(x\right)$ and
$p^{\prime\prime}\left(x\right)$ be its first and second derivatives,
respectively. Suppose that $p\left(x\right)$ is unimodal and $x_{m}$
denotes the mode defined by 
\[
x_{m}\in\underset{x}{\arg\max}\,p\left(x\right),
\]
so that $p^{\prime}\left(x_{m}\right)=0$ and $\left(x-x_{m}\right)p^{\prime}\left(x\right)\leq0$.
The concept of the \textit{flat-topped} $p\left(x\right)$ is illustrated
in Figure~\ref{fig:typical-flat-topped}, where $a,b\in\mathbb{R}$
are location parameters indicating the boundaries of the main part
and $a<x_{m}<b$. If $a$ and $b$ satisfy 
\[
\int_{-\infty}^{a}p\left(x\right)dx=\int_{a}^{x_{m}}\left(p\left(x_{m}\right)-p\left(x\right)\right)dx,
\]
\[
\int_{x_{m}}^{b}\left(p\left(x_{m}\right)-p\left(x\right)\right)dx=\int_{b}^{\infty}p\left(x\right)dx,
\]
then we have 
\[
a=x_{m}-\frac{1}{p\left(x_{m}\right)}\int_{-\infty}^{x_{m}}p\left(x\right)dx\qquad\mathrm{and}\qquad b=x_{m}+\frac{1}{p\left(x_{m}\right)}\int_{x_{m}}^{\infty}p\left(x\right)dx,
\]
which implies $p\left(x_{m}\right)=\left(b-a\right)^{-1}$. The interval
$\left[a,b\right]$ is expected, but not required, to be close to
the full width at half maximum (FWHM), i.e., $p\left(a\right)\approx p\left(x_{m}\right)/2\approx p\left(b\right)$.
Within its middle part $\left[x_{1},x_{2}\right]$ such that $x_{m}\in\left[x_{1},x_{2}\right]\subset\left(a,b\right)$,
we assume $p\left(x\right)$ is nearly constant. Then let us say that
$p$ is $\left(\varDelta,\varepsilon\right)$-\textit{flat-topped}
if for a given $\varepsilon>0$ there exists $\varDelta=x_{2}-x_{1}>0$
that satisfies 
\begin{equation}
1-\frac{1}{p\left(x_{m}\right)\varDelta}\int_{x_{1}}^{x_{1}+\varDelta}p\left(x\right)dx<\varepsilon.\label{ineq:flat-cond-delta}
\end{equation}
If $p\left(x\right)$ is concave within $\left[x_{1},x_{2}\right]$,
this condition may be substituted by 
\begin{equation}
1-\frac{p\left(x_{1}\right)+p\left(x_{2}\right)}{2p\left(x_{m}\right)}<\varepsilon.\label{ineq:flat-cond-x1x2}
\end{equation}
For example, the PDF of $\mathcal{U}\left(a,b\right)$ defined by
\begin{equation}
p_{U}^{}\left(x\mid a,b\right)=\begin{cases}
\frac{1}{b-a} & \textrm{for }a\leq x\leq b\\
0 & \textrm{otherwise}
\end{cases}\label{eq:u-def}
\end{equation}
is $\left(\varDelta,\varepsilon\right)$-\textit{flat-topped} for
any $\varepsilon>0$ and $0<\varDelta<b-a$. Therefore, any PDF that
approaches $p_{U}^{}\left(x\mid a,b\right)$ can be $\left(\varDelta,\varepsilon\right)$-\textit{flat-topped}
if close enough. Of course, $p\left(x\right)$ is required neither
to be \textit{smooth} (class $C^{\infty}$) nor \textit{flat} (all
derivatives vanish at $x\in\left[x_{1},x_{2}\right]$) in a calculus
sense.

In practice, however, a rigorous evaluation of $\left(\varDelta,\varepsilon\right)$
is not necessary. Alternatively, without using $\varDelta$, we simply
say that $p$ is $\varepsilon$-\textit{flat-topped} if for a given
$\varepsilon>0$ there exist $a$ and $b$ such that 
\begin{equation}
\left|p^{\prime\prime}\left(x_{m}\right)\right|\left|\frac{a-b}{p^{\prime}\left(a\right)-p^{\prime}\left(b\right)}\right|<\varepsilon.\label{ineq:flat-cond-d2pm}
\end{equation}
In the following sections, this inequality is mainly used for estimating
parameters that determine the \textit{flat-topped} shape, though there
is no clear boundary between \textit{flat-topped} and bell-shaped,
even if $\varepsilon\ll1$. 

%fig:typical-flat-topped

% subsection{Generalized normal distribution}
\subsection{Generalized normal distribution}

The generalized normal (or exponential power) distribution~\cite{Varanasi,Crooks,Liu}
is defined by the following PDF for all $x\in\mathbb{R}$: 
\begin{equation}
p_{GN}^{}\left(x\mid\mu,s,\beta\right)=\frac{\beta}{2s\varGamma\left(1/\beta\right)}\exp\left(-\left|\frac{x-\mu}{s}\right|^{\beta}\right),\label{eq:pgn}
\end{equation}
where $\mu\in\mathbb{R}$ is a location parameter, $s>0$ is a scale
parameter, $\beta>0$ is a shape parameter, and $\Gamma$ denotes
the gamma function. This includes the normal distribution $\mathcal{N}\left(\mu,\sigma^{2}\right)$,
which is given by 
\begin{equation}
p^{}_{N}\left(x\mid\mu,\sigma^{2}\right)=p_{GN}^{}\left(x\mid\mu,\sqrt{2}\sigma,2\right).\label{eq:pn-def}
\end{equation}
The cumulative distribution function (CDF) is expressed as 
\begin{align*}
P_{GN}\left(x\mid\mu,s,\beta\right) & =\int_{-\infty}^{x}p_{GN}^{}\left(y\mid\mu,s,\beta\right)dy\\
 & =\frac{1}{2}+\frac{\mathrm{sgn}\left(x-\mu\right)}{2\varGamma\left(1/\beta\right)}\,\gamma\!\left(\frac{1}{\beta},\left|\frac{x-\mu}{s}\right|^{\beta}\right)
\end{align*}
where $\mathrm{sgn\left(\cdot\right)}$ denotes the sign function
and $\mathrm{\gamma}\left(\cdot,\cdot\right)$ denotes the lower incomplete
gamma function.

Let $\mu_{f}\left(n\right)$ denote the $n$-th central moment of
a function $f$. If $n$ is even, then 
\[
\mu_{p_{GN}^{}}\left(n\right)=\frac{s^{n}\Gamma\left(\frac{n+1}{\beta}\right)}{\Gamma\left(\frac{1}{\beta}\right)}=\frac{s^{n}\Gamma\left(\frac{n+1}{\beta}+1\right)}{\left(n+1\right)\Gamma\left(\frac{1}{\beta}+1\right)}
\]
where we have used $\Gamma\left(x+1\right)=x\Gamma\left(x\right)$,
and so the kurtosis $\kappa$ is given by 
\begin{equation}
\kappa_{p_{GN}^{}}=\frac{\mu_{p_{GN}^{}}\left(4\right)}{\mu_{p_{GN}^{}}\left(2\right)^{2}}=\frac{9\,\Gamma\left(5/\beta+1\right)\Gamma\left(1/\beta+1\right)}{5\,\Gamma\left(3/\beta+1\right)^{2}}.\label{eq:kurtosis-pgn}
\end{equation}
Hence, $\lim_{\beta\rightarrow\infty}\kappa_{p_{GN}^{}}=9/5$, which is equal to the kurtosis of $p_{U}^{}$.

\begin{figure}[h]
\begin{centering}
 \includegraphics[scale=1.0]{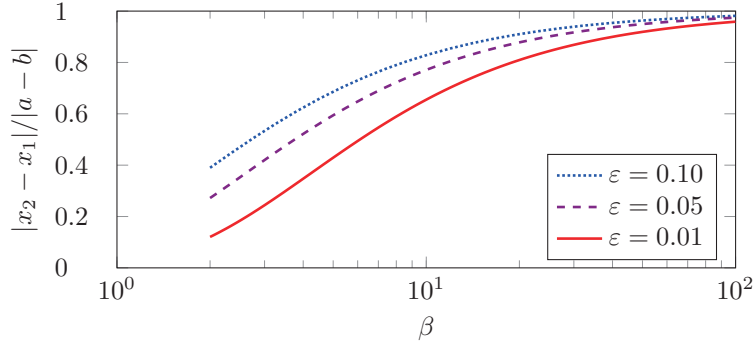}
\caption{The ratio of the \textit{flat-topped} interval $\left|x_{2}-x_{1}\right|$
to $\left|a-b\right|$ }
\label{fig:ratio-of-flat-topped}
\end{centering}
\end{figure}

The condition~(\ref{ineq:flat-cond-d2pm}) for \textit{flat-topped}
shape is checked as follows. The first and second derivatives of $p_{GN}^{}\left(x\right)$ for $x\neq\mu$ are 
\[
p_{GN}^{\prime}\left(x\mid\mu,s,\beta\right)=-\frac{\mathrm{sgn}\left(x-\mu\right)\beta^{2}}{2s^{2}\varGamma\left(1/\beta\right)}\left|\frac{x-\mu}{s}\right|^{\beta-1}\exp\left(-\left|\frac{x-\mu}{s}\right|^{\beta}\right)
\]
\[
p_{GN}^{\prime\prime}\left(x\mid\mu,s,\beta\right)=-\frac{\beta^{2}\left(\frac{\beta-1}{s}-\frac{\beta}{s}\left|\frac{x-\mu}{s}\right|^{\beta}\right)\left|\frac{x-\mu}{s}\right|^{\beta-2}\exp\left(-\left|\frac{x-\mu}{s}\right|^{\beta}\right)}{2s^{2}\varGamma\left(1/\beta\right)}.
\]
If $a=\mu-s\left(\ln2\right)^{1/\beta}$ and $b=\mu+s\left(\ln2\right)^{1/\beta}$,
then 
\[
p_{GN}^{}\left(x\mid\mu,s,\beta\right)=\frac{\beta\left(\ln2\right)^{1/\beta}}{\left(b-a\right)\varGamma\left(1/\beta\right)}2^{-\left|\frac{2x-a-b}{b-a}\right|^{\beta}}
\]
\[
\frac{p_{GN}^{\prime}\left(a\right)-p_{GN}^{\prime}\left(b\right)}{a-b}=-\frac{\beta^{2}}{4s^{3}\varGamma\left(1/\beta\right)}\left(\ln2\right)^{1-2/\beta}.
\]
The mode is at $x_{m}=\mu$ so that we have 
\[
p_{GN}^{\prime\prime}\left(\mu\right)=\begin{cases}
-\frac{2}{\sqrt{\pi}s^{3}} & \textrm{if }\beta=2,\\
0 & \textrm{if }\beta>2.
\end{cases}
\]
Therefore, $p_{GN}^{}$ is \textit{$\varepsilon$-flat-topped} for any
$\varepsilon>0$ if $\beta>2$. The interval $\left[x_{1},x_{2}\right]$
of $x$ that satisfies the condition $1-p_{GN}^{}\left(x\right)/p_{GN}^{}\left(\mu\right)\leq\varepsilon$
increases with increasing $\beta$ according to the equation 
\[
\left|\frac{x_{1}-x_{2}}{a-b}\right|=\left|\log_{2}\left(1-\varepsilon\right)\right|^{1/\beta}
\]
as shown in Fig.~\ref{fig:ratio-of-flat-topped}. This is consistent
with the characteristics of $\kappa_{p_{GN}^{}}$ in~(\ref{eq:kurtosis-pgn})
and $\lim_{\beta\rightarrow\infty}p_{GN}^{}\left(x\mid\mu,s,\beta\right)=p_{U}^{}\left(x\mid\mu-s,\mu+s\right)$
pointwise. Thus $\beta$ is the dominant parameter controlling the
\textit{flat-topped} shape.

% fig:ratio-of-flat-topped

Unfortunately, it is not easy to estimate $\beta$ using ML estimation.
Let $X=\left\{ X_{1},\ldots,X_{N}\right\} $ be an independent and
identically distributed (i.i.d.) sample with density $p_{GN}^{}$. The
log-likelihood is given by 
\begin{align*}
l_{GN}\left(\mu,s,\beta\mid X\right) & =\sum_{i}^{N}\ln p_{GN}^{}\left(X_{i}\mid\mu,s,\beta\right)\\
 & =-N\ln\left(2s\varGamma\left(\frac{1}{\beta}+1\right)\right)-\sum_{i}^{N}\left|\frac{X_{i}-\mu}{s}\right|^{\beta}.
\end{align*}
The ML estimator $\left(\hat{\mu},\hat{s},\hat{\beta}\right)=\arg\max_{\mu,s,\beta}\left\{ l_{GN}\right\} $
has no closed-form solution. It is necessary to use iterative numerical
optimization algorithms to obtain an approximate solution,
as discussed in~\cite{Varanasi,Liu}. In addition, another disadvantage
of $p_{GN}^{}$ is that it is not suitable for modeling asymmetrically
distributed data. 

% section{General Forms of Flat-topped PDFs}
\section{General Forms of Flat-topped PDFs}

There can be four types of \textit{flat-topped} PDF. Type~A is a
compound distribution, and Type~B is its modified version to fit
asymmetric distributions. Type~C is a generalization of the Cauchy
distribution and provides a new multivariate elliptical PDF. Type~D
is a function obtained by flattening the peak of arbitrary unimodal
functions. These types are not disjoint, and some PDFs may belong
to more than one type.

% \subsection{Type A: Compounding with uniform distribution}
\subsection{Type A: Compounding with uniform distribution}

The PDF of Type~A is defined by 
\begin{align}
p_{A}^{}\left(x\mid a,b,s\right) & =\int_{-\infty}^{\infty}f\left(x;u,s\right)p_{U}^{}\left(u\mid a,b\right)du\nonumber \\
 & =\frac{1}{b-a}\left(F\left(x;a,s\right)-F\left(x;b,s\right)\right)\label{eq:pa-def}
\end{align}
where $a$ and $b$ are location parameters, $s$ is a scale parameter, $f\left(x;u,s\right)$
is a continuous unimodal PDF with location parameter $u$, and $F\left(x;u,s\right)$
is its CDF such that 
\begin{equation}
F\left(x;u,s\right)=\int_{-\infty}^{x}f\left(y;u,s\right)dy=\int_{-\infty}^{x}f\left(\frac{y-u}{s};0,1\right)\frac{dy}{s}.\label{eq:F-def}
\end{equation}
The shape of $p_{A}^{}$ varies with parameters, from bell-shaped to
rectangular. The change of parameters
\begin{equation}
m=\frac{a+b}{2}\hspace{20bp}\mathrm{and}\hspace{20bp}r=\frac{b-a}{2}\label{eq:m_r}
\end{equation}
yields another integral form 
\[
p_{A}^{}\left(x\mid a,b,s\right)=p_{A}^{}\left(x\mid m-r,m+r,s\right)=\frac{1}{2r}\int_{\frac{x-m}{s}-\frac{r}{s}}^{\frac{x-m}{s}+\frac{r}{s}}f\left(y;0,1\right)dy
\]
so that we have 
\[
\lim_{r/s\rightarrow0}p_{A}^{}\left(x\mid a,b,s\right)=\frac{1}{s}f\left(\frac{x-m}{s};0,1\right)=f\left(x;m,s\right).
\]
On the other hand, if $F$ approaches a step function, then $p_{A}^{}$
approaches $p_{U}^{}$. To be more precise, let $P_{A}$ and $P_{U}$
be the CDFs of $p_{A}^{}$ and $p_{U}^{}$, respectively. If $F$ has the
property that $\lim_{s\rightarrow0}F\left(x;u,s\right)=H\left(x-u\right)$
where $H\left(\cdot\right)$ denotes the Heaviside step function,
then it follows that $\lim_{s\rightarrow0}P_{A}\left(x\mid a,b,s\right)=P_{U}\left(x\mid a,b\right)$.
This implies $p_{A}^{}\left(x\mid a,b,s\right)$ approaches $p_{U}^{}\left(x\mid a,b\right)$
(hereafter abbreviated as $p_{A}^{}\rightarrow p_{U}^{}$) as $s\rightarrow0$.
The distributional derivative of $H\left(x\right)$ is the Dirac delta
function $\delta\left(x\right)$ so that this property can be expressed
in the form 
\[
\lim_{s\rightarrow0}p_{A}^{}\left(x\mid a,b,s\right)=\int_{-\infty}^{\infty}p_{U}^{}\left(y\mid a,b\right)\delta\left(x-y\right)dy=p_{U}^{}\left(x\mid a,b\right).
\]
In general, if a distribution obtained by compounding $f\left(\mathbf{x}\right)$
with $g\left(\mathbf{x}\right)$ for $\mathbf{x}\in\mathbb{R}^{n}$
has the form of the convolution $p\left(\mathbf{x}\right)=\int_{\mathbb{R}^{n}}f\left(\mathbf{x}-\mathbf{y}\right)g\left(\mathbf{y}\right)d\mathbf{y}$,
then it can be expected that $p\left(\mathbf{x}\right)\rightarrow f\left(\mathbf{x}\right)$
as $g\left(\mathbf{y}\right)\rightarrow\delta\left(\mathbf{y}\right)$
and $p\left(\mathbf{x}\right)\rightarrow g\left(\mathbf{x}\right)$
as $f\left(\mathbf{y}\right)\rightarrow\delta\left(\mathbf{y}\right)$. 

If $m$ is the mean of $p_{A}^{}$, then the kurtosis $\kappa$ of $p_{A}^{}\left(x\mid m-r,m+r,s\right)$
is given by 
\begin{equation}
\kappa_{\,p_{A}^{}}=\frac{\mu_{f}\left(4\right)+2\mu_{f}\left(2\right)\left(\frac{r}{s}\right)^{2}+\frac{1}{5}\left(\frac{r}{s}\right)^{4}}{\left(\mu_{f}\left(2\right)+\frac{1}{3}\left(\frac{r}{s}\right)^{2}\right)^{2}}\label{eq:k-of-typeA}
\end{equation}
where $\mu_{f}\left(n\right)$ denotes the $n$-th central moment
of $f\left(x;0,1\right)$ (see Appendix A). If $\mu_{f}\left(4\right)$
and $\mu_{f}\left(2\right)$ are finite, then $\lim_{r/s\rightarrow0}\kappa_{p_{A}^{}}=\kappa_{f}$
and $\lim_{r/s\rightarrow\infty}\kappa_{p_{A}^{}}=\kappa_{p_{U}^{}}=9/5$.
There are a variety of PDFs available for $f$, such as normal ($\kappa=3$),
logistic ($\kappa=4.2$), Laplace ($\kappa=6$), Student's t ($\kappa=3+6/\left(\nu-4\right)$) and Cauchy distributions ($\kappa$
is undefined). If $f$ is leptokurtic, then $p_{A}^{}$ includes a variety of PDFs ranging from leptokurtic
to platykurtic. 

If $f$ is symmetric about the mode $m$ and satisfies the condition~(\ref{ineq:flat-cond-d2pm}), i.e., 
\[
\frac{2r}{s}\frac{f^{\prime}\left(-r/s;0,1\right)}{\left(f\left(0;0,1\right)-f\left(-2r/s;0,1\right)\right)}<\varepsilon,
\]
then $p_{A}^{}$ is \textit{$\varepsilon$-flat-topped}. Considering
that $p_{A}^{}$ approaches $p_{U}^{}$ as $s\rightarrow0$, this condition
can be simplified in the form $s/r<\varepsilon_{s}$ or $1-F\left(m;a,s\right)+F\left(m;b,s\right)<\varepsilon_{F}$.

% subsection{Type B: Product of sigmoid functions}
\subsection{Type B: Product of sigmoid functions}

The PDF of Type~B is defined by 
\begin{equation}
p_{B}^{}\left(x\mid a,b,s,t\right)=c\,F\left(x;a,s\right)\left(1-G\left(x;b,t\right)\right)\label{eq:pb-def}
\end{equation}
where $F$ and $G$ are sigmoid functions given by~(\ref{eq:F-def})
in Type~A and $c>0$ is a normalizing constant. In much the same
way as Type~A, $p_{B}^{}$ approaches $p_{U}^{}$ as $s,t\rightarrow0$,
so that it must be \textit{flat-topped} when $s$ and $t$ are small
enough. The advantage of Type~B is that $s$ and $t$ are independently
modifiable to fit the lower and upper tails to asymmetrically distributed
data. A disadvantage is that the dependence of $c$ on the parameters
cannot generally be expressed in closed form. However, if $F^{\prime}\left(x;0,1\right)$
and $G^{\prime}\left(x;0,1\right)$ are symmetric about zero and $p_{B}$
is \textit{flat-topped}, i.e., $1-F\left(x_{m};a,s\right)+G\left(x_{m};b,t\right)<\varepsilon_{F}\ll1$,
then $c$ can be approximated as $c\approx\left(b-a\right)^{-1}$
as shown later. In that case, the ML estimation becomes simpler. 

% subsection{Type C: Generalization of Cauchy distribution}
\subsection{Type C: Generalization of Cauchy distribution}

The third type can be expressed as 
\begin{equation}
p_{C}^{}\left(x\right)=\frac{c}{h+g\left(x\right)}\label{eq:pc-def}
\end{equation}
where $h>0$ is a positive constant and $g\left(x\right)$ is a non-negative
continuous U-shaped function that satisfies $\lim_{x\rightarrow\pm\infty}g\left(x\right)=\infty$.
Furthermore, $g\left(x\right)$ is supposed to satisfy $\left(x-x_{m}\right)g^{\prime}\left(x\right)\geq0$,
$0\leq g\left(x_{m}\right)\ll h$ for the mode $x_{m}\in\arg\min_{x}g\left(x\right)$,
and $g\left(a\right)=h=g\left(b\right)$ for the location parameters
$a<x_{m}<b$. The condition (\ref{ineq:flat-cond-d2pm})
is rewritten as 
\[
\frac{\left|g^{\prime\prime}\left(x_{m}\right)\right|}{\left(h+g\left(x_{m}\right)\right)^{2}}\left|\frac{4h^{2}\left(a-b\right)}{g^{\prime}\left(a\right)-g^{\prime}\left(b\right)}\right|<\varepsilon.
\]
This type is so fundamental that it includes the Cauchy (Lorentz)
distribution and can be easily extended to multivariate distributions. 

% subsection{Type D: Peak flattening}
\subsection{Type D: Peak flattening}

The PDF of this type is obtained by flattening the peak of bell-shaped
functions, which can be expressed as 
\begin{equation}
p_{D}^{}\left(x\right)=c\,\varPsi\left(\alpha\,f\left(x\right)\right)\label{eq:pd-def}
\end{equation}
where $c>0$ is a normalizing constant, $\alpha>0$ is a shape parameter,
$f\left(x\right)>0$ is a unimodal function of $x$, and $\varPsi\left(x\right)$
is a concave function satisfying $\varPsi\left(0\right)=0$, $\varPsi\left(x\right)\geq0$,
$\varPsi^{\prime}\left(x\right)\geq0$, and $\varPsi^{\prime\prime}\left(x\right)<0$
for $x\geq0$. If $\varPsi$ is a ``saturation'' function such that
$\lim_{x\rightarrow\infty}\varPsi^{\prime}\left(x\right)=0$, it is
easy to flatten the peak of $f\left(x\right)$. The advantage of this
type is that the \textit{flat-topped} shape can be applied to various unimodal function $f$, even if it has heavy tails. If $\varPsi$ is invertible, any \textit{flat-topped} PDF can be expressed in this form, though it might be more complicated.

% section{Specific Examples of Flat-topped PDFs}
\section{Specific Examples of Flat-topped PDFs}

This section presents some computationally tractable examples of the
\textit{flat-topped} PDFs. It seems that the Gaussian is typical for
$f$; However, the logistic function is advantageous for $F$, as
shown in 4.2 and 4.3. Its successful extensions for multivariate elliptical
distributions are proposed in 4.5.

% subsection{Uniform Gaussian Mixture}
\subsection{Uniform Gaussian Mixture}

The most likely component of the compound distribution of Type~A is
the normal distribution. Let $f_{N}$ and $F_{N}$ be the PDF and
CDF of $\mathcal{N}\left(m,\sigma^{2}\right)$, respectively, i.e.,
$f_{N}\left(x;u,s\right)=p^{}_{N}\left(x\mid u,s^{2}\right)$ and 
\[
F_{N}\left(x;u,s\right)=\frac{1}{2}\left(1+\mathrm{erf}\left(\frac{x-u}{\sqrt{2}s}\right)\right)
\]
where $\mathrm{erf}(\cdot)$ is the error function. It follows from~(\ref{eq:pa-def}) that we have 
\begin{equation}
p_{AN}^{}\left(x\mid a,b,s\right)=\frac{1}{2\left(b-a\right)}\left(\mathrm{erf}\left(\frac{x-a}{\sqrt{2}s}\right)-\mathrm{erf}\left(\frac{x-b}{\sqrt{2}s}\right)\right).\label{eq:pan-def}
\end{equation}
According to~(\ref{ineq:flat-cond-d2pm}),\textit{ $p_{AN}^{}\left(x\mid a,b,s\right)$}
is\textit{ $\varepsilon$-flat-topped} if 
\[
\left|\frac{\left(a-b\right)\,p_{AN}^{\prime\prime}\left(x_{m}\right)}{p_{AN}^{\prime}\left(a\right)-p_{AN}^{\prime}\left(b\right)}\right|<\frac{2\left(\frac{a-b}{2s}\right)^{2}}{\mathrm{\exp}\left(\frac{1}{2}\left(\frac{a-b}{2s}\right)^{2}\right)-1}<\varepsilon.
\]
The CDF of $p_{AN}^{}$ is expressed as 
\begin{align*}
P_{AN}\left(x\right)= & \frac{1}{2}+\frac{s}{2\left(b-a\right)}\left[\left(\frac{x-a}{s}\right)\mathrm{erf}\left(\frac{x-a}{\sqrt{2}s}\right)+\sqrt{\frac{2}{\pi}}\mathrm{\exp}\left(-\left(\frac{x-a}{\sqrt{2}s}\right)^{2}\right)\right.\\
 & \hspace{56bp}\left.-\left(\frac{x-b}{s}\right)\mathrm{erf}\left(\frac{x-b}{\sqrt{2}s}\right)-\sqrt{\frac{2}{\pi}}\mathrm{\exp}\left(-\left(\frac{x-b}{\sqrt{2}s}\right)^{2}\right)\right].
\end{align*}
A drawback is that the non-elementary function $\mathrm{erf}(\cdot)$
makes calculations somewhat intractable.

% subsection{Symmetric Type A using Logistic Function}
\subsection{Symmetric Type A using Logistic Function}

The most useful function for $F$ must be a logistic function such
that 
\begin{equation}
F_{L}\left(x;a,s\right)=\frac{1}{1+\exp\left(\frac{a-x}{s}\right)}=\frac{1}{2}\left(1+\mathrm{\tanh}\left(\frac{x-a}{2s}\right)\right).\label{eq:F-logi-def}
\end{equation}
The PDF $p_{AL}^{}\left(x\right)$ for $x\in\mathbb{R}$ is expressed
as 
\begin{align}
p_{AL}^{}\left(x\mid a,b,s\right) & =\frac{1}{b-a}\left(\frac{1}{1+\exp\left(\frac{a-x}{s}\right)}-\frac{1}{1+\exp\left(\frac{b-x}{s}\right)}\right)\nonumber \\
 & =\frac{1-\exp\left(\frac{a-b}{s}\right)}{\left(b-a\right)\left(1+\exp\left(\frac{a-x}{s}\right)\right)\left(1+\exp\left(\frac{x-b}{s}\right)\right)}\nonumber \\
 & =\frac{1}{2r}\left(\frac{\sinh\left(\frac{r}{s}\right)}{\cosh\left(\frac{x-m}{s}\right)+\cosh\left(\frac{r}{s}\right)}\right)
\label{eq:pal-def}
\end{align}
where $m=\left(a+b\right)/2$ and $r=\left(b-a\right)/2$. The first
equation analogous to a simple neural network model has been applied
to a mixture model~\cite{Alivanoglou}. The second equation shows
that $p_{AL}^{}\left(x\right)$ is a special case of the Perks distribution~\cite{Perks}
and also belongs to types B and C. The CDF is expressed as 
\begin{align*}
P_{AL}\left(x\right) & =\frac{s}{b-a}\ln\left(\frac{1+\exp\left(\frac{x-a}{s}\right)}{1+\exp\left(\frac{x-b}{s}\right)}\right)\\
 & =\frac{1}{2}+\frac{s}{r}\mathrm{artanh}\left(\tanh\left(\frac{x-m}{2s}\right)\tanh\left(\frac{r}{2s}\right)\right).
\end{align*}
The inverse cumulative distribution function (quantile function) is
\begin{align*}
P_{AL}^{-1}\left(v\right) & =a+s\ln\left(\frac{1-\exp\left(\frac{a-b}{s}v\right)}{\exp\left(\frac{a-b}{s}v\right)-\exp\left(\frac{a-b}{s}\right)}\right)\\
 & =m+2s\,\mathrm{artanh}\left(\tanh\left(\frac{r}{s}\left(v-\frac{1}{2}\right)\right)\coth\left(\frac{r}{2s}\right)\right)
\end{align*}
for a probability $v\in\left(0,1\right)$.

%fig:pal
\begin{figure}[h]
\begin{centering}
\includegraphics[scale=0.9]{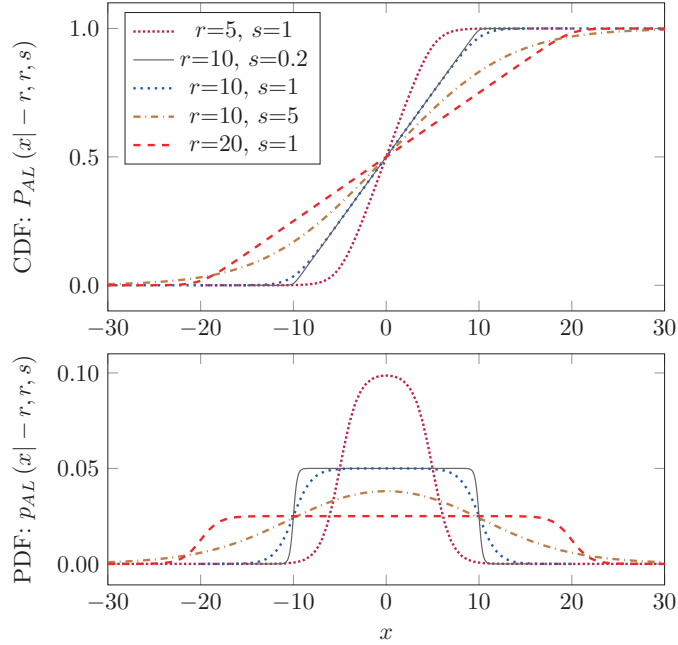}
\caption{Plots of $p_{AL}^{}$ and $P_{AL}$ for different parameter values.}
\label{fig:pal}
\end{centering}
\end{figure}

The shapes of $p_{AL}^{}$ and $P_{AL}$ depends on $r$ and $s$, as
shown in Figure~\ref{fig:pal}. Since $p_{AL}^{}\left(x\right)$ is
symmetric about $x_{m}=m$, the mean is $m$ and the skewness is zero.
The kurtosis is given by 
\[
\kappa_{p_{AL}^{}}=\frac{9}{5}+\frac{12}{5\left(1+\left(\frac{r}{\pi s}\right)^{2}\right)}
\]
which follows from~(\ref{eq:k-of-typeA}), so that $1.8<\kappa_{p_{AL}^{}}\leq4.2$
(see Appendix~B). If $r/s=\pi$,
then $\kappa=3$, i.e., the excess kurtosis $\kappa-3=0$ (mesokurtic),
which is similar to a normal distribution. In practice, however, a
better approximation to $\mathcal{N}\left(0,1\right)$ is given by
$p_{AL}^{}\left(x\mid-r_{\!\scriptscriptstyle N},r_{\!\scriptscriptstyle N},s_{\!\scriptscriptstyle N}\right)$ where $r_{\!\scriptscriptstyle N}=\sqrt{\ln4}-0.2\approx0.97741$
and $s_{\!\scriptscriptstyle N}=r_{\!\scriptscriptstyle N}/\pi+0.166\approx0.47712$, for which the error is estimated
as 
\[
\left|p^{}_{N}\left(x\mid0,1\right)-p_{AL}^{}\left(x\mid-r_{\!\scriptscriptstyle N},r_{\!\scriptscriptstyle N},s_{\!\scriptscriptstyle N}\right)\right|<0.0043.
\]
Hence, $\mathcal{N}\left(\mu,\sigma^{2}\right)$ can be approximated
as 
\begin{equation}
p^{}_{N}\left(x\mid\mu,\sigma^{2}\right)\approx p_{AL}^{}\left(x\mid\mu\!-\!\sigma\,r_{\!\scriptscriptstyle N},\,\mu\!+\!\sigma\,r_{\!\scriptscriptstyle N},\,\sigma\,s_{\!\scriptscriptstyle N}\right),\label{eq:pal-app-pn}
\end{equation}
though it is slightly leptokurtic ($\kappa\approx3.48$). 

According to~(\ref{ineq:flat-cond-d2pm}), $p_{AL}^{}\left(x\mid m-r,m+r,s\right)$
is \textit{$\varepsilon$-flat-topped} if 
\[
\left(4r/s\right)\mathrm{csch}\left(r/s\right)<\varepsilon
\]
(see Appendix C). Under this condition, $p_{AL}^{}$ can
be substituted for the above-mentioned $p_{AN}^{}$ given by~(\ref{eq:pan-def}), and vice versa, by using the following approximation:
\begin{equation}
p_{AN}^{}\left(x\mid a,b,s\right)\approx p_{AL}^{}\left(x\mid a,b,0.5877\,s\right).\label{eq:pan-app-pal}
\end{equation}
This is because $F_{N}\left(x;0,1\right)$ can be approximated by
$F_{L}\left(x;0,0.5877\right)$~\cite{Bowling}, where $\left|F_{N}\left(x;0,1\right)-F_{L}\left(x;0,0.5877\right)\right|<0.01$.
If $p_{AL}^{}$ is \textit{flat-topped}, the effect of the difference is limited to near the boundaries $a$ and $b$ so that it may be less influential in ML estimation. 

% subsection{Asymmetric Type B using Logistic Function}
\subsection{Asymmetric Type B using Logistic Function}

Type~B is useful for adjusting asymmetric tails to skewed data distributions.
In the following PDF, for example, its lower and upper tails mainly
depend on $s$ and $t$, respectively:  
\begin{align}
p_{BL}^{}\left(x\mid a,b,s,t\right) & =cF_{L}\left(x;a,s\right)\left(1-F_{L}\left(x;b,t\right)\right)\nonumber \\
 & =\frac{c}{\left(1+\exp\left(\frac{a-x}{s}\right)\right)\left(1+\exp\left(\frac{x-b}{t}\right)\right)}\label{eq:pbl-def}
\end{align}
where we have used $F_{L}\left(-x;-b,t\right)=1-F_{L}\left(x;b,t\right)$.
This $p_{BL}^{}\left(x\mid a,b,s,t\right)$ is \textit{$\varepsilon$-flat-topped}
if 
\[
6\left(\frac{b-a}{s}\coth\left(\frac{b-a}{2s}\right)+\frac{b-a}{t}\coth\left(\frac{b-a}{2t}\right)\right)\exp\left(\frac{a-b}{s+t}\right)<\varepsilon
\]
(see Appendix D). Unfortunately, $p_{BL}^{}$ cannot be integrated in closed form for $s\neq t$.
In practice, however, if $p_{BL}^{}$ is \textit{flat-topped}, then $c\approx1/\left(b-a\right)$
and $p_{BL}^{}$ can be approximated simply as 
\begin{equation}
p_{BL}^{}\left(x\mid a,b,s,t\right)\approx\begin{cases}
\frac{1}{b-a}F_{L}\left(x;a,s\right) & x<x_{m},\\
\frac{1}{b-a}\left(1-F_{L}\left(x;b,t\right)\right) & x\geq x_{m},
\end{cases}\label{eq:pbl-flat-approx}
\end{equation}
which is convenient to estimate the parameters (see Appendix~E). 

The second-best $F$ may be the CDF of the double exponential (Laplace)
distribution defined by 
\[
F_{D}\left(x;a,s\right)=\begin{cases}
\frac{1}{2}\exp\left(\frac{x-a}{s}\right) & x<a,\\
1-\frac{1}{2}\exp\left(-\frac{x-a}{s}\right) & x\geq a,
\end{cases}
\]
and then we have 
\begin{equation}
p_{BD}^{}\left(x\mid a,b,s,t\right)=cF_{D}\left(x;a,s\right)\left(1-F_{D}\left(x;b,t\right)\right)\label{eq:pbd-def}
\end{equation}
where 
\[
c=\left(b-a+\left(\frac{s^{2}}{s^{2}-t^{2}}\right)\frac{s}{2}\exp\left(\frac{a-b}{s}\right)+\left(\frac{t^{2}}{t^{2}-s^{2}}\right)\frac{t}{2}\exp\left(\frac{a-b}{t}\right)\right)^{-1}.
\]
Thus, $p_{BD}^{}$ can be integrated in closed form so that the approximation
\begin{equation}
p_{BL}^{}\left(x\mid a,b,s,t\right)\approx p_{BD}^{}\left(x\mid a,b,s\ln4,t\ln4\right)\label{eq:pbl-app-pbd}
\end{equation}
is useful for understanding the properties of $p_{BL}^{}$.

% subsection{Asymmetric Type A}
\subsection{Asymmetric Type A}

The PDF of Type~A can be asymmetric depending on the asymmetry of
$F$ such as a CDF of the skew normal distribution~\cite{OHagan}.
From the viewpoint of tractability, it is better to use elementary
functions, for example: 
\[
F_{LS}\left(x;a,s,\lambda\right)=\frac{1}{2}\left[1+\tanh\left(\frac{x-a}{2s}+\lambda\left[\sqrt{\left(\frac{x-a}{2s}\right)^{2}+1}-1\right]\right)\right]
\]
where $\lambda\in\left(-1,1\right)$ is a skew parameter. This parameterization
is designed to satisfy $F_{LS}\left(x;a,s,0\right)=F_{L}\left(x;a,s\right)$,
$F_{LS}\left(a\right)=1/2$, and $F_{LS}^{\prime}\left(a\right)=1/\left(4s\right)$.
As $\lambda$ increases, the lower tail becomes heavier and the upper
tail becomes lighter. However, the lower and upper tails cannot be
adjusted independently, which is less convenient than the above-mentioned
Type~B. 

% subsection{Symmetric Type C}
\subsection{Symmetric Type C}

This section shows three examples of $g\left(x\right)$ in~(\ref{eq:pc-def}):
$y^{\beta}$, $\exp\left(y^{\beta}\right)$, and $\cosh\left(y^{\beta}\right)$
where $y=\left|x-m\right|/s$, $m\in\mathbb{R}$, and $s,\beta>0$. 

First, a special case of the generalized Cauchy (generalized Pearson
VII) distribution~\cite{Crooks,Rider,Pearson} is defined by 
\begin{equation}
p_{CC}^{}\left(x\mid m,s,\beta\right)=\frac{\beta}{2s\,B\left(1-\frac{1}{\beta},\frac{1}{\beta}\right)\left[1+\left|\frac{x-m}{s}\right|^{\beta}\right]}\label{eq:pc-power}
\end{equation}
where $B\left(1-1/\beta,1/\beta\right)=\pi/\sin\left(\pi/\beta\right)$
is the beta function. If $\beta>2$, then $p_{CC}^{}$ is \textit{$\varepsilon$-flat-topped}
for any $\varepsilon>0$. If $\beta=4$, it is called the Laha distribution~\cite{Laha}. For $\beta=6$, the kurtosis of $p_{CC}^{}\left(x\mid0,1,6\right)$
is 4 so that it is leptokurtic. A disadvantage is that the shape parameter
$\beta$ is difficult to estimate, as in $p_{GN}^{}$. 

Second, a new PDF is defined by 
\begin{equation}
p_{CF}^{}\left(x\mid m,r,s,\beta\right)=\frac{\beta}{2s\Gamma\left(\frac{1}{\beta}\right)F_{\frac{1}{\beta}-1}\left(\frac{r^{\beta}}{s^{\beta}}\right)\left[1+\exp\left(\frac{\left|x-m\right|^{\beta}-r^{\beta}}{s^{\beta}}\right)\right]}\label{eq:pc-fermi}
\end{equation}
where $r>0$ and $F_{j}\left(x\right)$ is the complete Fermi-Dirac
integral~\cite{Dingle} (see Appendix~F). Note that $F_{j}\left(x\right)$
has a numerical subscript (not to be confused with $F$ having a capital
letter subscript used for types A and B). We call $p_{CF}^{}$ a generalized
Fermi-Dirac distribution, though the Fermi function of the Fermi-Dirac
statistics in physics is not a PDF. The kurtosis of $p_{CF}^{}$ is given
by 
\[
\kappa_{\,p_{CF}^{}}=\frac{\Gamma\left(1/\beta\right)F_{1/\beta-1}\left(r^{\beta}/s^{\beta}\right)\Gamma\left(5/\beta\right)F_{5/\beta-1}\left(r^{\beta}/s^{\beta}\right)}{\left[\Gamma\left(3/\beta\right)F_{3/\beta-1}\left(r^{\beta}/s^{\beta}\right)\right]^{2}}.
\]
The special case of $\beta=1$ is a variant of the Fermi function
with normalizing constant: 
\[
p_{CF}^{}\left(x\mid m,r,s,1\right)=\frac{1}{2s\ln\left(1+\exp\left(\frac{r}{s}\right)\right)\left[1+\exp\left(\frac{\left|x-m\right|-r}{s}\right)\right]}.
\]
The\textit{ }shape of this PDF varies with $r$ and $s$, not with
$\beta$, which is much better than the generalized Gaussian $p_{GN}^{}$.
According to the condition~(\ref{ineq:flat-cond-x1x2}) for $x_{1}=m-r/2$
and $x_{2}=m+r/2$, if $\exp\left(-r/\left(2s\right)\right)<\varepsilon$,
then $p_{CF}^{}\left(x\mid m,r,s,1\right)$ is \textit{$\left(r,\varepsilon\right)$-flat-topped}
and similar but not superior to $p_{AL}^{}$. The special case of $\beta=2$
is a Ferreri distribution~\cite{Ferreri} and rewritten using parameters
$a$ and $b$ instead of $m$ and $r$ as 
\begin{equation}
p_{CE}^{}\left(x\mid a,b,s\right)=\frac{1}{\sqrt{\pi}s\,F_{-1/2}\left(r^{2}/s^{2}\right)\left[1+\exp\left(\frac{\left(x-a\right)\left(x-b\right)}{s^{2}}\right)\right]}.\label{eq:pce-def}
\end{equation}
This seems to be simple with respect to $x$, but unfortunately it
cannot be integrated in closed form. This $p_{CE}^{}$ is \textit{$\varepsilon$-flat-topped}
if $\mathrm{sech}^{2}\left(-\left(\frac{b-a}{2s}\right)^{2}\right)<\varepsilon$.
If $\beta>2$, then $p_{CF}^{}$ is \textit{$\varepsilon$-flat-topped}
for any $\varepsilon>0$. 

For multivariate elliptical distributions of $n$ dimensional vectors,
using the Mahalanobis distance
\begin{equation}
d_{M}\left(\mathbf{x},\mathbf{m},\boldsymbol{\Sigma}\right)=\left(\left(\mathbf{x}-\mathbf{m}\right)^{\mathsf{T}}\boldsymbol{\Sigma}^{-1}\left(\mathbf{x}-\mathbf{m}\right)\right)^{1/2}\label{eq:def-mahalanobis}
\end{equation}
where $\mathbf{x},\mathbf{m}\in\mathbb{R}^{n}$ and $\boldsymbol{\Sigma}$
is an $n\times n$ positive-definite matrix like a covariance matrix,
$p_{CF}^{}$ can be extended to be 
\begin{equation}
p_{CM}^{}\left(\mathbf{x}\mid\mathbf{m},\boldsymbol{\Sigma},r,t\right)=\frac{c_{M}}{1+\exp\left(\left[d_{M}\left(\mathbf{x},\mathbf{m},\boldsymbol{\Sigma}\right)^{n}-r^{n}\right]t\right)}\label{eq:pcm-def}
\end{equation}
where $r$ is a dispersion parameter, $t=1/s^{n}$ is a shape parameter
used for adjusting the slope of boundaries, and $c_{M}$ is a normalizing
constant given by 
\[
c_{M}=\frac{t\,\Gamma\left(n/2+1\right)}{\pi^{n/2}\ln\left(1+\exp\left(r^{n}t\right)\right)\left|\boldsymbol{\Sigma}\right|^{1/2}}
\]
(see Appendix G). A similar PDF has been proposed by Gasparini and
Ma~\cite{Gasparini}: the multivariate Fermi-Dirac distribution defined
as the form of 
\[
f\left(\mathbf{x}\right)=\frac{c_{G}}{1+\exp\left(\alpha+\lambda^{2}\left(\alpha\right)\left(\mathbf{x}-\mathbf{m}\right)^{\mathsf{T}}\boldsymbol{\Sigma}^{-1}\left(\mathbf{x}-\mathbf{m}\right)\right)}
\]
where $\lambda^{2}\left(\alpha\right)=F_{1/2}\left(\alpha\right)/F_{-1/2}\left(\alpha\right)$
and the normalizing constant is given by 
\[
c_{G}=\frac{\Gamma\left(n/2\right)\lambda^{n}\left(\alpha\right)}{\pi^{n/2}F_{n/2-1}\left(\alpha\right)\left|\boldsymbol{\Sigma}\right|^{1/2}}.
\]
This is disadvantageous because the dependence of $c$ on $\alpha$
cannot be expressed in closed form except for $n=2$.

\begin{figure}[h]
\begin{centering}
\includegraphics[width=0.9\textwidth]{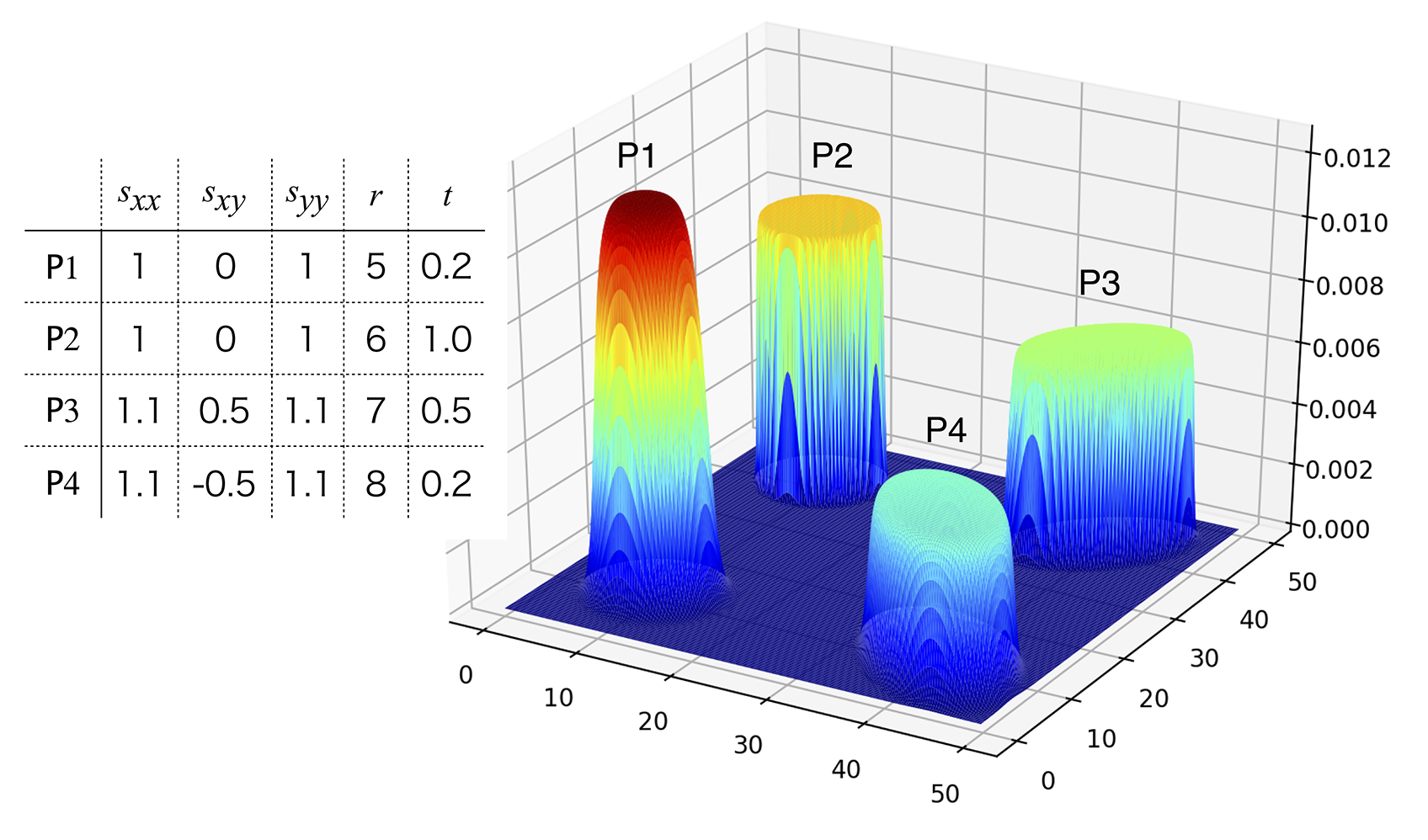}
\caption{Multivariate flat-topped distributions: $p_{CL}^{}$ ($n=2$) for different parameter values.}
\label{fig:pcl}
\end{centering}
\end{figure}

Third, one more new PDF is defined by 
\begin{equation}
p_{CH}^{}\left(x\mid m,r,s,\beta\right)=\frac{c_{H}\sinh\left(r^{\beta}/s^{\beta}\right)}{\cosh\left(\left|x-m\right|^{\beta}/s^{\beta}\right)+\cosh\left(r^{\beta}/s^{\beta}\right)}\label{eq:pch-def}
\end{equation}
where $r,s,\beta>0$ and 
\[
c_{H}=\frac{\beta}{2s\Gamma\left(1/\beta\right)\left[F_{1/\beta-1}\left(r^{\beta}/s^{\beta}\right)-F_{1/\beta-1}\left(-r^{\beta}/s^{\beta}\right)\right]}.
\]
The kurtosis of $p_{CH}^{}$ is given by 
\[
\kappa_{\,p_{CH}^{}}=\frac{\Gamma\left(\frac{1}{\beta}\right)\left[F_{\frac{1}{\beta}-1}\left(\frac{r^{\beta}}{s^{\beta}}\right)-F_{\frac{1}{\beta}-1}\left(-\frac{r^{\beta}}{s^{\beta}}\right)\right]\Gamma\left(\frac{5}{\beta}\right)\left[F_{\frac{5}{\beta}-1}\left(\frac{r^{\beta}}{s^{\beta}}\right)-F_{\frac{5}{\beta}-1}\left(-\frac{r^{\beta}}{s^{\beta}}\right)\right]}{\Gamma\left(\frac{3}{\beta}\right)^{2}\left[F_{\frac{3}{\beta}-1}\left(\frac{r^{\beta}}{s^{\beta}}\right)-F_{\frac{3}{\beta}-1}\left(-\frac{r^{\beta}}{s^{\beta}}\right)\right]^{2}}.
\]
The special case of $\beta=1$ is identical to $p_{AL}^{}$ defined by~(\ref{eq:pal-def}).
If $\beta\geq2$, then $p_{CH}^{}$ is always \textit{flat-topped}. For
multivariate elliptical distributions, in much the same way as $p_{CM}^{}$,
we have 
\begin{equation}
p_{CL}^{}\left(\mathbf{x}\mid\mathbf{m},\boldsymbol{\Sigma},r,t\right)=\frac{c_{L}\sinh\left(r^{n}t\right)}{\cosh\left(d_{M}\left(\mathbf{x},\mathbf{m},\boldsymbol{\Sigma}\right)^{n}t\right)+\cosh\left(r^{n}t\right)}\label{eq:pcl-def}
\end{equation}
where 
\[
c_{L}=\frac{\Gamma\left(n/2+1\right)}{\pi^{n/2}r^{n}\left|\boldsymbol{\Sigma}\right|^{1/2}}.
\]
This is very natural because $\pi^{n/2}r^{n}/\Gamma\left(n/2+1\right)$
corresponds to the volume of an n-dimensional ball of radius $r$. The
examples of 2-dimensional $p_{CL}^{}$ for various values of $\left\{ \boldsymbol{\Sigma},r,t\right\} $
are shown in Fig.~\ref{fig:pcl}. The parameters $\boldsymbol{\Sigma}$ and
$t=1/s^{n}$ are redundant so that we can impose a constraint on them
such as $\left|\boldsymbol{\Sigma}\right|=1$, and so the number of
necessary parameters is $\left(n+1\right)\left(n+2\right)/2$. However,
redundant parameters are helpful for quickly finding solutions to
parameter optimization problems. The log-likelihood equations for
$p_{CM}$ and $p_{CL}$ are expressed by elementary functions of their
parameters, which seems relatively simple in this type. 

% fig:pcl

% subsection{Symmetric Type D}
\subsection{Symmetric Type D}

There can be a variety of PDFs given by~(\ref{eq:pd-def}). The typical
examples of $\varPsi\left(x\right)$ are ``saturation'' functions
such as $1-\exp\left(-x\right)$, $\tanh\left(x\right)$, and $\arctan\left(x\right)$.
The unimodal functions for $f\left(x;m,s\right)$ can be not only
the PDFs of bell-shaped distributions but also simpler functions like
$y^{-\beta}$ and $\exp\left(-y^{\beta}\right)$ where $y=\left|x-m\right|/s$
and $\beta>0$. For example, a simple one is 
\begin{equation}
p_{DE}^{}\left(x\mid m,s\right)=\frac{1}{2\sqrt{\pi}s}\left(1-\exp\left\{ -\left(\frac{x-m}{s}\right)^{-2}\right\} \right)\label{eq:pde-def}
\end{equation}
which is \textit{$\varepsilon$-flat-topped} for any $\varepsilon>0$
with heavy tails. The CDF is expressed as 
\begin{align*}
P_{DE}\left(x\mid m,s\right) & =\frac{1}{2}\left[1+\left(\frac{x-m}{\sqrt{\pi}s}\right)\left(1-\exp\left\{ -\left(\frac{s}{x-m}\right)^{2}\right\} \right)\right.\\
 & \hspace{75bp}\left.+\mathrm{sgn}\left(x-m\right)-\mathrm{erf}\left(\frac{s}{x-m}\right)\right].
\end{align*}
Unfortunately, there are few other simple \textit{flat-topped} PDF
of this type, mainly because $c$ is rarely expressed in closed form.

% section{Maximum Likelihood Estimation}
\section{Maximum Likelihood Estimation}

This section discusses ML estimation for $p_{AL}^{}$, $p_{BL}^{}$, and
$p_{CL}^{}$. The likelihood equations of the \textit{flat-topped} PDFs
have no closed-form solution, so it is necessary to use iterative
methods to obtain approximate solutions.

%\subsection{Simplified gradient ascent}
\subsection{Simplified gradient ascent}

Let $X=\left\{ x_{1},\ldots,x_{N}\right\} $ be a given data set.
Assuming $X$ is i.i.d. with density $p_{AL}^{}\left(x\mid a,b,s\right)$
given by~(\ref{eq:pal-def}), the model parameters $a$, $b$, and $s$
are estimated by maximizing the log-likelihood function: 
\begin{equation}
l_{AL}\left(a,b,s;X\right)=\sum_{i=1}^{N}\ln p_{AL}^{}\left(x_{i}\mid a,b,s\right).\label{eq:lpal}
\end{equation}
Its partial derivatives are 
\begin{align*}
\partial_{a}l_{AL} & =\frac{1}{s}\sum_{i=1}^{N}\left(\frac{s}{b-a}-\frac{1}{\exp\left(\frac{b-a}{s}\right)-1}-1+F_{L}\left(x_{i};a,s\right)\right)\\
\partial_{b}l_{AL} & =-\frac{1}{s}\sum_{i=1}^{N}\left(\frac{s}{b-a}-\frac{1}{\exp\left(\frac{b-a}{s}\right)-1}-F_{L}\left(x_{i};b,s\right)\right)\\
\partial_{s}l_{AL} & =-\frac{1}{s^{2}}\sum_{i=1}^{N}\left(\frac{b-a}{\exp\left(\frac{b-a}{s}\right)-1}+\left(x_{i}-a\right)\left(1-F_{L}\left(x_{i};a,s\right)\right)\right.\\
 & \hspace{75bp}\hspace{75bp}\left.\vphantom{\frac{a}{\exp\left(\frac{b}{s}\right)}}-\left(x_{i}-b\right)F_{L}\left(x_{i};b,s\right)\right)
\end{align*}
(see Appendix~H). The likelihood equations obtained by setting these
derivatives equal to zero have no closed-form solution. However, we
can impose a constraint that
\[
\min\left\{ x_{i}\right\} <a<b<\max\left\{ x_{i}\right\} \qquad\mathrm{and}\qquad\frac{b-a}{4N}<s<\sigma,
\]
where $\sigma$ is the standard deviation of $X$. It seems to be
easy to find approximately optimal parameters by using iterative methods.
Although Newton's method is sometimes inappropriate for the case of
ill-conditioned Hessians, a simplified procedure that modifies each
parameter one by one as in a coordinate descent algorithm~\cite{Wright}
works well in practice. An example of iteration is as follows: 
\begin{align*}
a^{new} & =a^{old}+\eta_{a}\,\partial_{a}l_{AL}\left(a^{old},b^{old},s^{old};X\right)\\
b^{new} & =b^{old}+\eta_{b}\,\partial_{b}l_{AL}\left(a^{new},b^{old},s^{old};X\right)\\
s^{new} & =s^{old}+\eta_{s}\,\partial_{s}l_{AL}\left(a^{new},b^{new},s^{old};X\right)
\end{align*}
where $\eta_{a}$, $\eta_{b}$, and $\eta_{s}$ are coefficients for
step-size control such as $\eta_{a}\propto\left|\partial_{aa}l_{AL}\right|^{-1}$.
Metaheuristic optimization techniques using adaptive step-size control
are also applicable. 

For $p_{BL}^{}$ given by~(\ref{eq:pbl-def}), if it is \textit{flat-topped},
then the partial derivatives of its log-likelihood $l_{BL}$ can be
approximated as: 
\[
\partial_{a}l_{BL}\approx\frac{N}{b-a}-\frac{1}{s}\sum_{i=1}^{N}\left(1-F_{L}\left(x_{i};a,s\right)\right),\quad\partial_{b}l_{BL}\approx-\frac{N}{b-a}+\frac{1}{t}\sum_{i=1}^{N}F_{L}\left(x_{i};b,t\right),
\]
\[
\partial_{s}l_{BL}\approx\frac{1}{s^{2}}\sum_{i=1}^{N}\left(a-x_{i}\right)\left(1-F_{L}\left(x_{i};a,s\right)\right),\quad\partial_{t}l_{BL}\approx\frac{1}{t^{2}}\sum_{i=1}^{N}\left(x_{i}-b\right)F_{L}\left(x_{i};b,t\right).
\]
These approximations may be relatively easier than those of the other
\textit{flat-topped} PDFs. 

For $p_{CL}^{}$ given by~(\ref{eq:pcl-def}), the log-likelihood is 
\[
l_{CL}=N\ln\left(\frac{\Gamma\left(n/2+1\right)\sinh\left(r^{n}t\right)}{\pi^{n/2}r^{n}\left|\Sigma\right|^{1/2}}\right)-\sum_{i=1}^{N}\ln\left(\cosh\left(\rho_{i}^{n}t\right)+\cosh\left(r^{n}t\right)\right)
\]
where $\rho_{i}=d_{M}\left(\mathbf{x}_{i},\mathbf{m},\boldsymbol{\Sigma}\right)$
is given by (\ref{eq:def-mahalanobis}). Thus, the partial derivatives
of $l_{CL}$ with respect to parameters $\left\{ \mathbf{m},\boldsymbol{\Sigma}^{-1},r^{n},t\right\} $
can be evaluated using elementary functions. For instance, 
\[
\partial_{\boldsymbol{\Sigma}^{-1}}\,l_{CL}=\frac{N}{2}\mathbf{\boldsymbol{\Sigma}}-\frac{1}{2}\sum_{i=1}^{N}\frac{\sinh\left(\rho_{i}^{n}t\right)n\rho_{i}^{n-2}t}{\cosh\left(\rho_{i}^{n}t\right)+\cosh\left(r^{n}t\right)}\left(\mathbf{x}_{i}-\mathbf{m}\right)\left(\mathbf{x}_{i}-\mathbf{m}\right)^{\mathsf{T}}.
\]
Unfortunately, the computational cost of matrix operations is expensive
for large $n$. It is important to simplify the model by decomposing
it into factorized PDFs of fewer variables to reduce the cost. 

% \subsection{Advantage of the \textit{flat-topped} PDF}
\subsection{Advantage of the \textit{flat-topped} PDF}

The \textit{flat-topped} PDF can be adapted to fit a variety of distribution
shapes ranging from bell-shaped to rectangular, and it brings about
the increase of the log-likelihood. For example, suppose $X$ is an
i.i.d. sample from the uniform distribution $\mathcal{U}\left(a,b\right)$.
If this data set is modeled by a normal distribution, using ML estimation,
the best fit $p_{N}^{\ast}$ is estimated to be $\mathcal{N}\left(m,r^{2}/3\right)$
where $m=\left(a+b\right)/2$ and $r=\left(b-a\right)/2$. Hence, 
the expected log-likelihood is 
\begin{align*}
\mathbb{E}\left[\ln p_{N}^{\ast}\left(x\mid m,r^{2}/3\right)\right] & =\int_{-\infty}^{\infty}p_{U}\left(x\mid a,b\right)\ln p_{N}^{\ast}\left(x\mid m,r^{2}/3\right)dx\\
 & =-\ln2r-\frac{1}{2}\ln\frac{\pi e}{6}.
\end{align*}
If it is exactly modeled by $\mathcal{U}\left(a,b\right)$, the expected
log-likelihood increases by the Kullback-Leibler (KL) divergence of
$\mathcal{U}\left(a,b\right)$ with respect to $\mathcal{N}\left(m,r^{2}/3\right)$,
i.e., 
\[
D_{KL}\left(p_{U}^{}\parallel p_{N}^{\ast}\right)=\frac{1}{2}\ln\frac{\pi e}{6}\approx0.176.
\]
Although this value appears small, it should not be neglected, because
the values of $\ln\left(p_{U}^{}\left(x\right)/p_{N}^{\ast}\left(x\right)\right)$
can be positive or negative and cancel each other out in averaging.
In fact, the $L_{1}$ distance between them is 
\begin{align*}
D_{L_{1}}\left(p_{U}^{},p_{N}^{\ast}\right) & =\int_{-\infty}^{\infty}\left|p_{U}^{}\left(x\mid a,b\right)-p_{N}^{\ast}\left(x\mid m,r^{2}/3\right)\right|dx\\
 & =2\left(1-\sqrt{\frac{1}{3}\ln\left(\frac{6}{\pi}\right)}+\mathrm{erf}\left(\sqrt{\frac{1}{2}\ln\left(\frac{6}{\pi}\right)}\right)-\mathrm{erf}\left(\sqrt{\frac{3}{2}}\right)\right)\\
 & \approx0.395.
\end{align*}
This is not trivial considering that $\sup\left\{ D_{L_{1}}\left(p_{U}^{},p_{N}^{\ast}\right)\right\} =2$.

In an $n$-dimensional space, let $p_{MU}^{}\left(\mathbf{x}\mid\mathbf{m},r\right)$
denote the PDF of a multivariate uniform distribution such that 
\begin{equation}
p_{MU}^{}\left(\mathbf{x}\mid\mathbf{m},r\right)=\lim_{t\rightarrow\infty}p_{CL}^{}\left(\mathbf{x}\mid\mathbf{m},\mathbf{I},r,t\right)=\begin{cases}
\frac{\Gamma\left(n/2+1\right)}{\pi^{n/2}r^{n}} & \mathrm{if}\;\left\Vert \mathbf{x}-\mathbf{m}\right\Vert _{2}\leq r\\
0 & \mathrm{otherwise}
\end{cases},\label{eq:pmu-def}
\end{equation}
where $\mathbf{x},\mathbf{m}\in\mathbb{R}^{n}$ are $n$-dimensional
vectors and $\mathbf{I}$ is the $n\times n$ identity matrix substituted
for $\boldsymbol{\Sigma}$ in~(\ref{eq:pcl-def}), then its best-fit
model using a multivariate normal distribution is given by $p_{MN}^{\ast}\left(\mathbf{x}\mid\mathbf{m},\hat{\boldsymbol{\Sigma}}\right)$
where the elements of $\hat{\boldsymbol{\Sigma}}$ are $\sigma_{ii}=r^{2}/\left(n+2\right)$
for $i=1,2,\ldots,n$ and $\sigma_{ij}=0$ for $i\neq j$. The KL
divergence of $p_{MU}^{}\left(\mathbf{x}\mid\mathbf{m},r\right)$ with
respect to $p_{MN}^{\ast}\left(\mathbf{x}\mid\mathbf{m},\hat{\boldsymbol{\Sigma}}\right)$
is calculated as 
\[
D_{KL}\left(p_{MU}^{}\parallel p_{MN}^{\ast}\right)=\ln\Gamma\left(\frac{n}{2}+1\right)-\frac{n}{2}\ln\left(\frac{n}{2}+1\right)+\frac{n}{2}
\]
and the $L_{1}$ distance between them is 
\[
D_{L_{1}}\left(p_{MU}^{},p_{MN}^{\ast}\right)=2\left(1-\chi_{n}^{n}-\frac{\Gamma\left(n/2,\left(n/2+1\right)\chi_{n}^{2}\right)-\Gamma\left(n/2,n/2+1\right)}{\Gamma\left(n/2\right)}\right)
\]
where 
\[
\chi_{n}=\sqrt{\frac{2}{n+2}\ln\frac{\left(n/2+1\right)^{n/2}}{\Gamma\left(n/2+1\right)}}
\]
and $\Gamma\left(\cdot,\cdot\right)$ is the incomplete gamma function
(see Appendix~I). If $n=2$, then $\chi_{2}^{2}=\ln\sqrt{2}$, $D_{KL}\left(p_{MU}^{}\parallel p_{MN}^{\ast}\right)=1-\ln2\approx0.307$,
and $D_{L_{1}}\left(p_{MU}^{},p_{MN}^{\ast}\right)=1-\ln2+2/e^{2}\approx0.578$
where we have used $\Gamma\left(1,x\right)=e^{-x}$. Thus, both $D_{KL}\left(p_{MU}^{}\parallel p_{MN}^{\ast}\right)$
and $D_{L_{1}}\left(p_{MU}^{},p_{MN}^{\ast}\right)$ monotonically increase
with $n$ so that the \textit{flat-topped }PDF is more effective in
high dimensional spaces. 

% \section{Mixture Models}
\section{Mixture Models}

The mixture of \textit{flat-topped }PDFs can be useful for improving the goodness of fit of the $\text{GMM}$. 

%\subsection{Outline of model fitting}
\subsection{Outline of model fitting}

A practical estimation procedure consists of three steps as follows: 
\begin{enumerate}
\item Create a finite GMM using the EM (or VB) algorithm.
\item Improve the model by replacing each Gaussian component with a symmetric
$p_{AL}^{}\left(x\mid a,b,s\right)$ and using a generalized EM algorithm~\cite{Dempster}.
\item If the optimized $p_{AL}^{}\left(x\mid a,b,s\right)$ is \textit{flat-topped},
replace it with an asymmetric $p_{BL}^{}\left(x\mid a,b,s,t\right)$ and optimize in the same way.
\end{enumerate}
It is possible to build a mixture model using only \textit{flat-topped
}PDFs from scratch. However, the GMM is easier to build first and
becomes a standard for comparison. The GMM can be smoothly transformed
into a mixture of \textit{flat-topped} PDFs and further optimized,
which increases the log-likelihood. Even though the log-likelihood
improvement may be small, it is important to understand the characteristics of
the boundary regions of subpopulations. Moreover, $p_{AL}^{}$ can be
replaced with $p_{AN}^{}$ that is a uniform Gaussian mixture given
by~(\ref{eq:pan-def}), or we can restore the previously optimized
GMM if it is reasonable. A similar approach can also be applied to
modeling multivariate elliptical distributions using $p_{CM}^{}$ or
$p_{CL}^{}$.

%\subsection{Mixture of \textit{flat-topped} distributions}
\subsection{Mixture of \textit{flat-topped} distributions}

We consider a mixture model of the form 
\[
p_{FM}^{}\left(x\mid\boldsymbol{\theta}\right)=\sum_{k=1}^{K}\pi_{k}\,p_{AL}^{}\left(x\mid a_{k},b_{k},s_{k}\right)
\]
where $K$ is the number of mixture components, $\boldsymbol{\theta}=\left\{ \pi_{k},a_{k},b_{k},s_{k}\mid k=1,\ldots,K\right\} $
denotes model parameters, and $\pi_{k}\in\left[0,1\right]$ is the
mixing coefficients. Since $\sum_{k=1}^{K}\pi_{k}=1$, the number
of free parameters is $4K-1$. There is no closed-form solution for
maximizing the likelihood $\prod_{i=1}^{N}p_{FM}\left(x_{i}\mid\boldsymbol{\theta}\right)$
for an i.i.d data set $\mathbf{X}=\left\{ x_{1},\ldots,x_{N}\right\} $.
However, by introducing a latent variable $\mathbf{Z}=\left\{ z_{i,k}\in\left\{ 0,1\right\} \mid\sum_{k=1}^{K}z_{i,k}=1\right\} $
and considering the problem of maximizing the likelihood for the complete
data set $\left\{ \mathbf{X},\mathbf{Z}\right\} $ such that 
\[
L\left(\boldsymbol{\theta};\mathbf{X},\mathbf{Z}\right)=p\left(\mathbf{X},\mathbf{Z}\mid\boldsymbol{\theta}\right)=\prod_{i=1}^{N}\prod_{k=1}^{K}\left[\pi_{k}\,p_{AL}^{}\left(x\mid a_{k},b_{k},s_{k}\right)\right]^{z_{i,k}},
\]
we can find an approximate solution using a generalized EM algorithm.
For example, 
\begin{enumerate}
\item Choose an initial setting for the parameters. If a GMM has been already
obtained, each Gaussian component can be replaced with $p_{AL}^{}\left(x\mid a_{k},b_{k},s_{k}\right)$
using~(\ref{eq:pal-app-pn}).
\item E-step: Evaluate the expected complete data log-likelihood given by
\[
Q=\sum_{i=1}^{N}\sum_{k=1}^{K}w_{i,k}\left[\ln\pi_{k}+\ln p_{AL}^{}\left(x_{i}\mid a_{k},b_{k},s_{k}\right)\right]
\]
where $w_{i,k}$ denotes the probability that component $k$ is responsible
for generating $x_{i}$. This probability can be estimated as a posterior
probability with respect to the latent variables using Bayes' theorem
as follows: 
\[
w_{i,k}=\mathbb{E}\left[z_{i,k}\right]=\frac{\pi_{k}\,p_{AL}^{}\left(x_{i}\mid a_{k},b_{k},s_{k}\right)}{\sum_{j=1}^{K}\pi_{j}\,p_{AL}^{}\left(x_{i}\mid a_{j},b_{j},s_{j}\right)}.
\]
\item M-step: Update the parameters to increase $Q$ using $w_{i,k}$ as
follows: 
\begin{align*}
\pi_{k}^{new} & =\frac{1}{N}\sum_{i=1}^{N}w_{i,k}\\
a_{k}^{new} & =a_{k}^{old}+\eta_{a}\left(a_{k}^{old}\right)\sum_{i=1}^{N}w_{i,k}\left[\frac{\partial}{\partial a_{k}}\ln p_{AL}^{}\left(x_{i}\mid a_{k}^{old},b_{k}^{old},s_{k}^{old}\right)\right]\\
b_{k}^{new} & =b_{k}^{old}+\eta_{s}\left(b_{k}^{old}\right)\sum_{i=1}^{N}w_{i,k}\left[\frac{\partial}{\partial b_{k}}\ln p_{AL}^{}\left(x_{i}\mid a_{k}^{new},b_{k}^{old},s_{k}^{old}\right)\right]\\
s_{k}^{new} & =s_{k}^{old}+\eta_{s}\left(s_{k}^{old}\right)\sum_{i=1}^{N}w_{i,k}\left[\frac{\partial}{\partial s_{k}}\ln p_{AL}^{}\left(x_{i}\mid a_{k}^{new},b_{k}^{new},s_{k}^{old}\right)\right]
\end{align*}
where $\eta_{\theta}\left(\theta_{k}\right)$ is a coefficient for
step-size control of $\theta_{k}$ such that 
\[
\eta_{\theta}\left(\theta_{k}\right)\propto\left|\frac{\partial^{2}}{\partial\theta_{k}^{2}}\sum_{i=1}^{N}w_{i,k}\ln p_{AL}^{}\left(x_{i}|\theta_{k}\right)\right|^{-1}.
\]
\item Repeat E- and M-steps until the estimates converge.
\end{enumerate}
The M-step is almost the same as a single iteration of the iterative
method described in Section~5.1, except the
log-likelihood has the coefficient $w_{i,k}$. If the optimized  $p_{AL}^{}$ is \textit{flat-topped}, it can be replaced with $p_{BL}^{}$ to further
improve the log-likelihood in much the same way.

% \section{Experiments}
\section{Experiments}

In this section, the usefulness of the \textit{flat-topped} PDF is
demonstrated with simulation examples.

% \subsection{ML estimation for univariate distribution}
\subsection{ML estimation for univariate distribution}

\begin{figure}[h]
\begin{centering}
\includegraphics[width=0.8\textwidth]{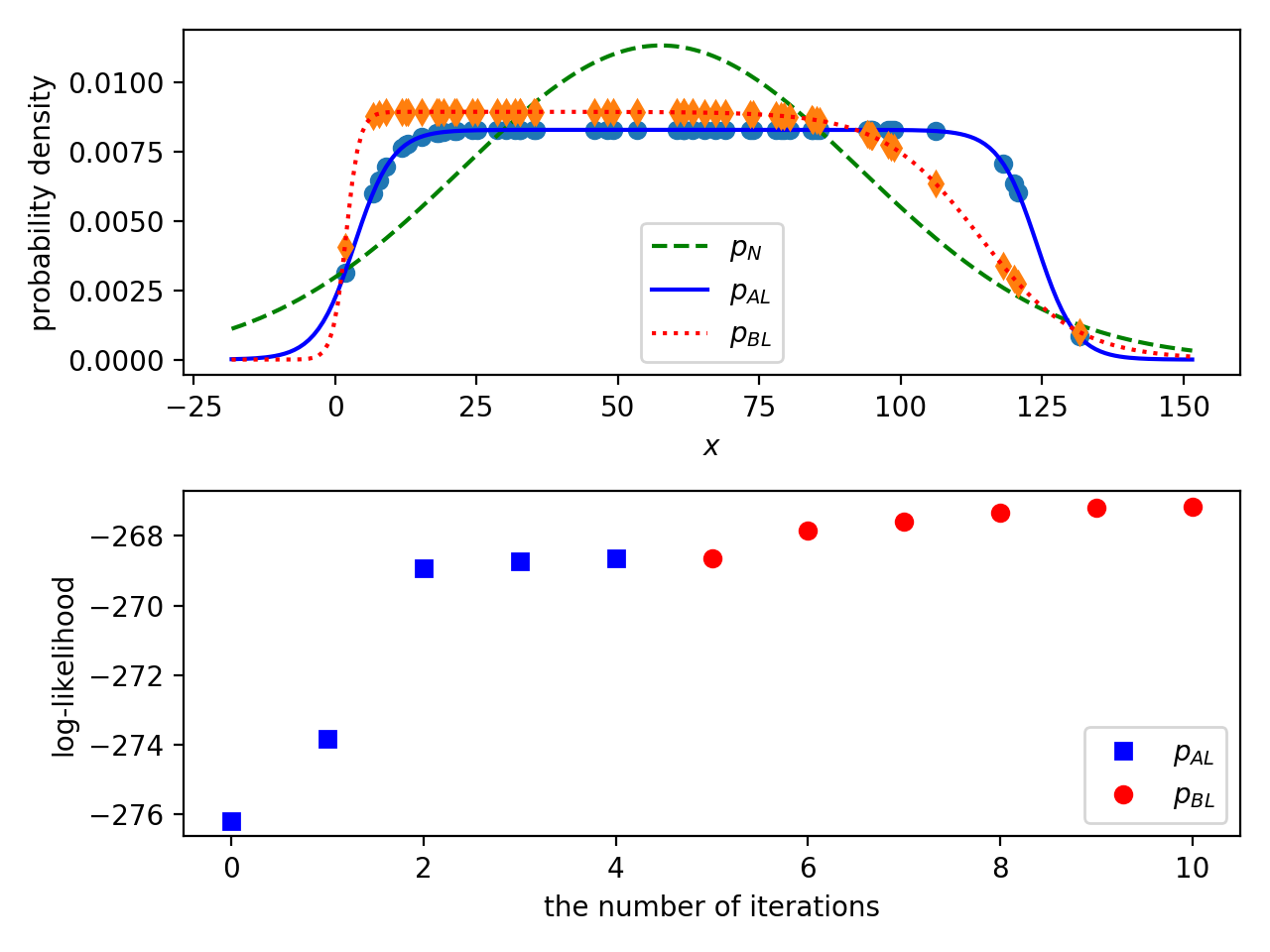}\caption{ML estimation for $p_{AL}^{}$ and $p_{BL}^{}$. Data points are marked
on each of the curves. }
\label{fig:ML-estimation} 
\end{centering}
\end{figure}

The iterative method for the ML estimation of $p_{AL}^{}$ described
in Section 5 generally works well. An example is illustrated in Fig.~\ref{fig:ML-estimation},
which shows the three PDFs $p_{N}^{}$, $p_{AL}^{}$, and $p_{BL}^{}$ fitted
for a test sample of $N=55$ data points: 40 are from $\mathcal{U}\left(0,100\right)$,
and 15 are from $\mathcal{N}\left(60,35^{2}\right)$. The parameters
of $p_{AL}^{}$ are initially set to approximate $p_{N}^{}$ using~(\ref{eq:pal-app-pn}).
The iterations almost converge rapidly and bring about a reasonable
increase in log-likelihood. If the sample $X$ is likely to be uniformly
distributed, it is advisable to start from $s=4s_{min}=\left(\max\left(X\right)-\min\left(X\right)\right)/N$,
$a=\min\left(X\right)+s$, and $b=\max\left(X\right)-s$, and keep
the constraint $s\geq s_{min}$ to avoid overflow and underflow. 

% fig:ML-estimation

% \subsection{Bivariate mixture modeling}
\subsection{Bivariate mixture modeling}

The advantage of the mixture model using the \textit{flat-topped}
PDFs (hereafter abbreviated FTM) over GMM is demonstrated in modeling
the following two-dimensional synthetic data. The focus is not only
on goodness of fit, as discussed in Section~5.2, but also on
parsimonious modeling based on AIC~\cite{Akaike} and BIC~\cite{Schwarz}.

% fig:3D-plots
\begin{figure}[h]
\begin{centering}
\noindent\begin{minipage}[t]{1\columnwidth}%
\includegraphics[width=1\textwidth]{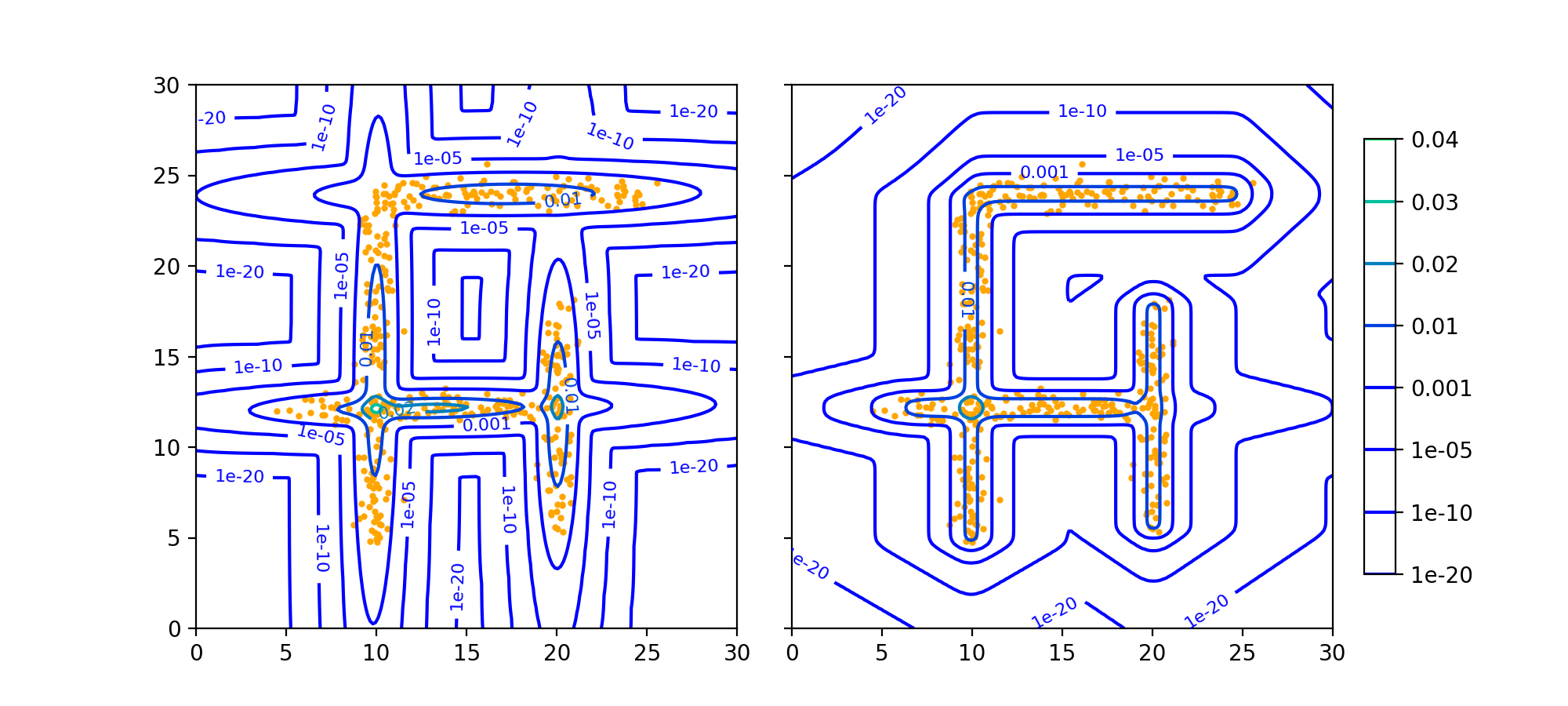}
\caption{Contours of the PDFs of GMM (left) and FTM (right) fitted for the
sample of 406 data points (orange dots). Both models consist of four
components $\left(K=4\right)$. }
\label{fig:Contours}
\end{minipage}

\noindent\begin{minipage}[t]{1\columnwidth}%
\includegraphics[width=1\textwidth]{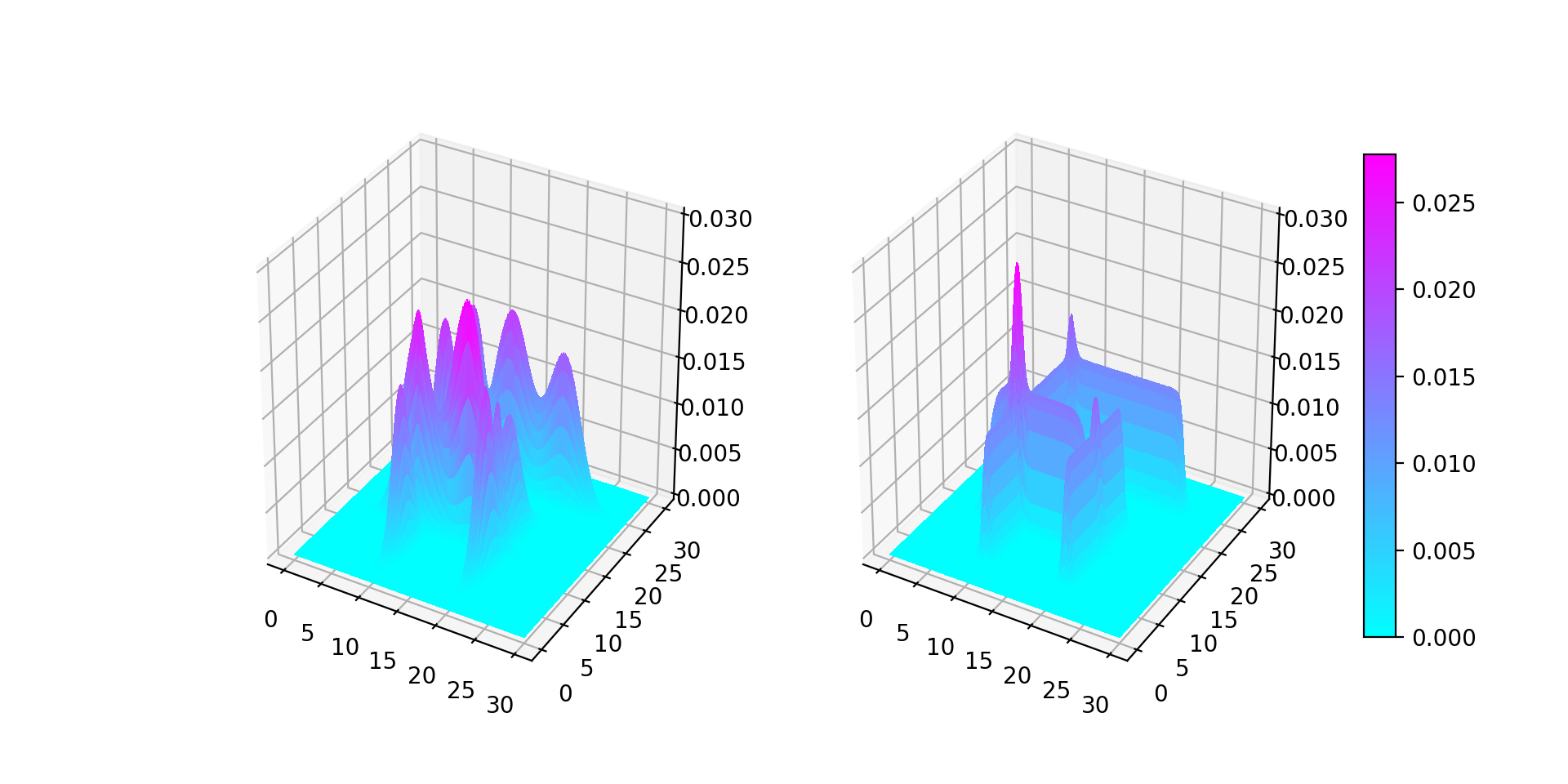}

\caption{The 3D plots of the PDFs of GMM (left: $K=9$) and FTM (right: $K=4$)
that minimize AIC.}
\label{fig:3D-plots}
\end{minipage}
\end{centering}
\end{figure}

The goodness of fit of the estimated models in the following simulations
can be qualitatively assessed by simply looking at PDF plots. The
data points are generated as $\mathbf{u}+\boldsymbol{\epsilon}$
where $\mathbf{u}$ is a two-dimensional vector representing a random
point from a uniform distribution on line segments in a plane and
$\boldsymbol{\epsilon}$ is a small isotropic Gaussian random vector
with mean zero. For the sake of convenience, the line segments are
aligned with the coordinate axes, and so the mixture components of
FTM can be modeled by $p_{FM}^{}\left(x,y\right)=p_{AL}^{}\left(x\right)p_{AL}^{}\left(y\right)$
where $\left(x,y\right)$ denotes Cartesian coordinates. Figure~\ref{fig:Contours}
shows the contours of two PDFs fitted for the data points $\left(N=427\right)$;
the left plot shows a GMM $\left(K=4\right)$ estimated using the
EM algorithm implemented in Scikit-learn~\cite{Scikit-learn} and
the right plot a FTM $\left(K=4\right)$ fitted using the generalized
EM algorithm presented in Section 6.3. Both models have the lowest
BIC values. Figure~\ref{fig:3D-plots} shows surface plots of the
PDFs of a GMM ($K=9$) and the same FTM $\left(K=4\right)$ that have
the lowest AIC values. The obvious disadvantages of the GMMs are excess
peaks, unreal tails, and unclear boundaries in this case.

% fig:AIC-and-BIC
\begin{figure}[h]
\begin{centering}
\includegraphics[scale=0.30]{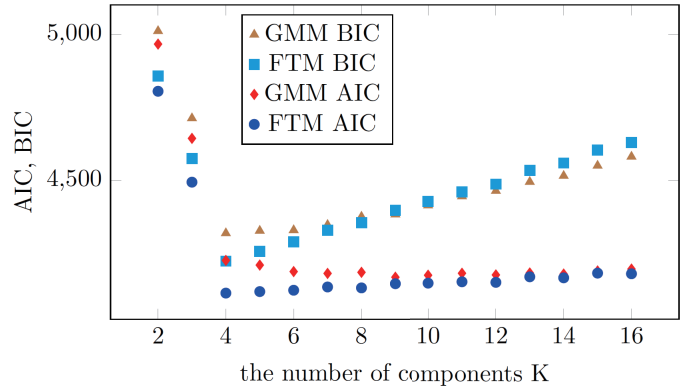}
\caption{AIC and BIC}
\label{fig:AIC-and-BIC}
\end{centering}
\end{figure}

% \begin{table}
\begin{table}[h]
\begin{centering}
\caption{The minimum values of AIC and BIC}
\label{tab:abic}
\begin{tabular}{|c|c|c|r|c|c|r@{\extracolsep{0pt}.}l|r|c|c|r@{\extracolsep{0pt}.}l|}
\hline 
\multirow{2}{*}{$X$} & \multirow{2}{*}{$N$} & \multirow{2}{*}{$K$} & \multicolumn{5}{c|}{GMM} & \multicolumn{5}{c|}{FTM}\tabularnewline
\cline{4-13} \cline{5-13} \cline{6-13} \cline{7-13} \cline{9-13} \cline{10-13} \cline{11-13} \cline{12-13} 
 &  &  & it  & $l$/$N$ & AIC & \multicolumn{2}{c|}{BIC} & it  & $l$/$N$ & AIC & \multicolumn{2}{c|}{BIC}\tabularnewline
\hline 
\hline 
\multirow{2}{*}{A} & \multirow{2}{*}{427} & 4 & 31 & -4.895 & 4227 & \multicolumn{2}{c|}{\textbf{\textcolor{red}{\emph{4320}}}} & 49 & -4.755 & \textbf{\textcolor{red}{\emph{4115}}} & \multicolumn{2}{c|}{\textbf{\textcolor{red}{\emph{4224}}}}\tabularnewline
\cline{3-13} \cline{4-13} \cline{5-13} \cline{6-13} \cline{7-13} \cline{9-13} \cline{10-13} \cline{11-13} \cline{12-13} 
 &  & 9 & 9 & -4.468 & \textbf{\textcolor{red}{\emph{4169}}} & \multicolumn{2}{c|}{4384} & 100 & -4.711 & 4147 & \multicolumn{2}{c|}{4398}\tabularnewline
\hline 
\multirow{3}{*}{B} & \multirow{3}{*}{1281} & 4 & 20 & -4.874 & 12534 & \multicolumn{2}{c|}{12652} & 61 & -4.772 & \textbf{\textcolor{red}{\emph{12280}}} & \multicolumn{2}{c|}{\textbf{\textcolor{red}{\emph{12419}}}}\tabularnewline
\cline{3-13} \cline{4-13} \cline{5-13} \cline{6-13} \cline{7-13} \cline{9-13} \cline{10-13} \cline{11-13} \cline{12-13} 
 &  & 5 & 11 & -4.854 & 12494 & \multicolumn{2}{c|}{\textbf{\textcolor{red}{\emph{12643}}}} & 62 & -4.771 & 12292 & \multicolumn{2}{c|}{12467}\tabularnewline
\cline{3-13} \cline{4-13} \cline{5-13} \cline{6-13} \cline{7-13} \cline{9-13} \cline{10-13} \cline{11-13} \cline{12-13} 
 &  & 15 & 25 & -4.745 & \textbf{\textcolor{red}{\emph{12334}}} & \multicolumn{2}{c|}{12793} & 300 & -4.726 & 12317 & \multicolumn{2}{c|}{12853}\tabularnewline
\hline 
\end{tabular}
\par\end{centering}
\centering{}(it: the number of iterations, $l/N$: average log-likelihood)
\end{table}

The difference between AIC and BIC is significant for model selection.
Figure~\ref{fig:AIC-and-BIC} shows the plot of AIC and BIC values,
where the numbers of free parameters in the two-dimensional GMM and
FTM are $6K-1$ and $7K-1$, respectively. Naturally, both the AIC
and BIC values of FTM are minimum at the number of the given line
segments $\left(K=4\right)$. On the other hand, for the optimal number
of GMM components, BIC indicates just the same $K=4$, but AIC suggests
$K=9$. The AIC values reflect subtle situation of model fitting.
Table~\ref{tab:abic} shows the lowest values (indicated by italics)
of AIC evaluated for the above models. It signifies that the number
of $K$ minimizing AIC for GMM increases with $N$. In other words,
the AIC values imply that the GMM ($K=4$) is insufficient for improving
goodness of fit and needs more Gaussian components, even though they
considerably overlap each other. That is quite reasonable, considering
that the optimal model is almost equivalent to an infinite uniform
mixture of Gaussians. The \textit{flat-topped} PDF can approximate
such a model using minimal parameters. 

In the basis function decomposition of an arbitrary PDF, it is essential
to choose appropriate basis functions. The FTM $\left(K=4\right)$
seems much better than the GMM ($K=9$) in goodness of fit in the
case of Figure~\ref{fig:3D-plots}. However, the log-likelihood values
in Table~\ref{tab:abic} indicate not much difference between them.
That implies the likelihood is not the best measure of goodness of
fit, and neither is KL divergence. It is desirable to develop another
criterion for model selection to compare a wider variety of models. 

% \section{Concluding Remarks}
\section{Concluding Remarks}

The most tractable univariate \textit{flat-topped} PDF is $p_{AL}^{}$ defined by~(\ref{eq:pal-def}). It is obtained by compounding a logistic distribution with a uniform distribution, and its shape varies
with its parameters, from bell-shaped to rectangular. For asymmetric
\textit{flat-topped} distributions, $p_{BL}^{}$ defined by~(\ref{eq:pbl-def})
is available. Furthermore, a generalized Fermi-Dirac distribution
$p_{CM}^{}$ defined by~(\ref{eq:pcm-def}) and its variant $p_{CL}^{}$
defined by~(\ref{eq:pcl-def}) are advantageous for modeling multivariate
elliptical distributions. Although there is no closed-form solution
for the ML estimates of model parameters, we can obtain approximate
solutions using iterative methods. Thus, they are useful as a component
of a mixture model that can be optimized using the generalized EM
algorithm. Even in GMM, if it contains some data points distributed
uniformly, it will be worthwhile to replace the Gaussians with \textit{flat-topped}
PDFs to improve goodness of fit and make the model as parsimonious
as possible. In such a situation, AIC values may suggest that the
Gaussian components are needed more than that indicated by BIC values.

% appendix A
\specialsection*{Appendix A. Kurtosis of $p_{A}^{}$}

The $n$-th central moment of $p_{A}^{}\left(x\mid-r,r,s\right)$ given by~(\ref{eq:pa-def}), for a positive even integer $n$,  is evaluated as 
\begin{align*}
\mu_{\,p_{A}^{}}\left(n\right) & =\int_{-\infty}^{\infty}x^{n}\left\{ \int_{-\infty}^{\infty}f\left(x;u,s\right)p_{U}^{}\left(u\mid-r,r\right)du\right\} dx\\
 & =\int_{-\infty}^{\infty}p_{U}^{}\left(u\mid-r,r\right)\left\{ \int_{-\infty}^{\infty}x^{n}f\left(\frac{x-u}{s};0,1\right)\frac{1}{s}dx\right\} du\\
 & =\frac{1}{2r}\int_{-r}^{r}\left\{ \int_{-\infty}^{\infty}\left(sy+u\right)^{n}f\left(y;0,1\right)dy\right\} du\\
 & =\frac{1}{2r}\int_{-r}^{r}\left\{ \int_{-\infty}^{\infty}\sum_{i=0}^{n}\left(\begin{array}{c}
n\\
i
\end{array}\right)\left(sy\right)^{n-i}u^{i}f\left(y;0,1\right)dy\right\} du\\
 & =\frac{1}{2r}\sum_{i=0}^{n}\left(\begin{array}{c}
n\\
i
\end{array}\right)s^{n-i}\mu_{f}\left(n-i\right)\frac{r^{i+1}-\left(-r\right)^{i+1}}{i+1}
\end{align*}
where $y=\left(x-u\right)/s$ and $\mu_{f}\left(n\right)$ denotes the
$n$-th central moment of $f\left(x;0,1\right)$. It follows that
\[
\mu_{\,p_{A}^{}}\left(2\right)=s^{2}\mu_{f}\left(2\right)+\frac{r^{2}}{3},
\]
\[
\mu_{\,p_{A}^{}}\left(4\right)=s^{4}\mu_{f}\left(4\right)+6s^{2}\mu_{f}\left(2\right)\frac{r^{2}}{3}+\frac{r^{4}}{5}.
\]
Thus, we have 
\begin{equation}
\kappa_{\,p_{A}^{}}=\frac{\mu_{\,p_{A}^{}}\left(4\right)}{\mu_{\,p_{A}^{}}\left(2\right)^{2}}=\frac{\mu_{f}\left(4\right)+2\mu_{f}\left(2\right)\left(\frac{r}{s}\right)^{2}+\frac{1}{5}\left(\frac{r}{s}\right)^{4}}{\left(\mu_{f}\left(2\right)+\frac{1}{3}\left(\frac{r}{s}\right)^{2}\right)^{2}}.\label{eq:A-k-typeA}
\end{equation}

% appendix B
\specialsection*{Appendix B. Central moments of $p_{AL}^{}$ }

The second and fourth central moments of $f_{L}\left(x;0,1\right)=F_{L}^{\prime}\left(x;0,1\right)=\mathrm{sech}^{2}\left(x/2\right)/4$
are evaluated as $\mu_{f_{L}}\left(2\right)=\pi^{2}/3$ and $\mu_{f_{L}}\left(4\right)=7\pi^{4}/15$,
respectively (see reference~\cite{Balakrishnan}). It follows from~(\ref{eq:A-k-typeA})
that we have 
\[
\kappa_{\,p_{AL}^{}}=\frac{9}{5}+\frac{12}{5\left(1+\left(\frac{r}{\pi s}\right)^{2}\right)}.
\]

Alternatively, the central moments of $p_{AL}^{}$ given by~(\ref{eq:pal-def})
can be evaluated directly using the complete Fermi\textendash Dirac
integral: 
\begin{equation}
F_{j}\left(x\right)=\frac{1}{\Gamma\left(j+1\right)}\int_{0}^{\infty}\frac{t^{j}}{e^{t-x}+1}dt=-\mathrm{Li}_{j+1}\left(-e^{x}\right)\label{eq:FD-integ}
\end{equation}
where $\mathrm{Li}_{n}$ is the polylogarithm function defined by
\[
\mathrm{Li}_{n}\left(z\right)=\sum_{k=1}^{\infty}\frac{z^{k}}{k^{n}}.
\]
Furthermore, $\mathrm{Li}_{n}$ satisfies the following relation~\cite{Lewin}
\[
\mathrm{Li}_{n}\left(-z\right)+\left(-1\right)^{n}\mathrm{Li}_{n}\left(-1/z\right)=-\frac{1}{n!}\left(\ln z\right)^{n}+2\sum_{k=1}^{\left\lfloor n/2\right\rfloor }\frac{\left(\ln z\right)^{n-2k}}{\left(n-2k\right)!}\mathrm{Li}_{2k}\left(-1\right)
\]
where $\left\lfloor x\right\rfloor $ denotes the greatest integer
less than or equal to $x$. Since $\mathrm{Li}_{2}\left(-1\right)=-\pi^{2}/12$
and $\mathrm{Li}_{4}\left(-1\right)=-7\pi^{4}/720$, we have 
\[
\mathrm{Li}_{3}\left(-z\right)-\mathrm{Li}_{3}\left(-1/z\right)=-\frac{1}{6}\left(\ln z\right)^{3}-\frac{\pi^{2}}{6}\ln z,
\]
\[
\mathrm{Li}_{5}\left(-z\right)-\mathrm{Li}_{5}\left(-1/z\right)=-\frac{1}{120}\left(\ln z\right)^{5}-\frac{\pi^{2}}{36}\left(\ln z\right)^{3}-\frac{7\pi^{4}}{360}\ln z.
\]
By using these relations, the $n$-th central moment $\mu_{\,p_{AL}^{}}\left(n\right)$ of $p_{AL}^{}$, for a positive even integer $n$, is expressed as 
\begin{align*}
\mu_{\,p_{AL}^{}}\left(n\right) & =\int_{-\infty}^{\infty}\frac{\left(x-m\right)^{n}}{b-a}\left(\frac{1}{1+\exp\left(\frac{a-x}{s}\right)}-\frac{1}{1+\exp\left(\frac{b-x}{s}\right)}\right)dx\\
 & =\frac{s^{n+1}}{r}\int_{0}^{\infty}\left(\frac{y^{n}}{1+\exp\left(y-\frac{r}{s}\right)}-\frac{y^{n}}{1+\exp\left(y+\frac{r}{s}\right)}\right)dy\\
 & =\frac{s^{n+1}}{r}\varGamma\left(n+1\right)\left\{ -\mathrm{Li}_{n+1}\left(-\exp\left(\frac{r}{s}\right)\right)+\mathrm{Li}_{n+1}\left(-\exp\left(-\frac{r}{s}\right)\right)\right\} 
\end{align*}
where $b-a=2r>0$, $m=\left(a+b\right)/2$, and $y=\left(m-x\right)/s$.
It follows that 
\[
\mu_{\,p_{AL}^{}}\left(2\right)=\frac{2s^{3}}{r}\left(\frac{1}{6}\left(\frac{r}{s}\right)^{3}+\frac{\pi^{2}}{6}\left(\frac{r}{s}\right)\right)=\frac{s^{2}\pi^{2}}{3}\left(\left(\frac{r}{\pi s}\right)^{2}+1\right)
\]
\begin{align*}
\mu_{\,p_{AL}^{}}\left(4\right) & =\frac{s^{5}}{r}4!\left(\frac{1}{120}\left(\frac{r}{s}\right)^{5}+\frac{\pi^{2}}{36}\left(\frac{r}{s}\right)^{3}+\frac{7\pi^{4}}{360}\left(\frac{r}{s}\right)\right)\\
 & =\frac{s^{4}\pi^{4}}{15}\left(3\left(\frac{r}{\pi s}\right)^{4}+10\left(\frac{r}{\pi s}\right)^{2}+7\right).
\end{align*}
Thus, we obtain the same result from $\kappa_{\,p_{AL}^{}}=\mu_{\,p_{AL}^{}}\left(4\right)/\mu_{\,p_{AL}^{}}\left(2\right)^{2}$
.

% appendix C
\specialsection*{Appendix C. Condition for flat-topped $p_{AL}^{}$}

The PDF $p_{AL}^{}\left(x\right)$ is given by
\[
p_{AL}^{}\left(x\right)=\frac{1}{2r}\left(\frac{\sinh\left(\frac{r}{s}\right)}{\cosh\left(\frac{x-m}{s}\right)+\cosh\left(\frac{r}{s}\right)}\right)
\]
in~(\ref{eq:pal-def}). Its first and second derivatives are 
\[
p_{AL}^{\prime}\left(x\right)=-\frac{1}{2rs}\left(\frac{\sinh\left(\frac{r}{s}\right)\sinh\left(\frac{x-m}{s}\right)}{\left(\cosh\left(\frac{x-m}{s}\right)+\cosh\left(\frac{r}{s}\right)\right)^{2}}\right),
\]
\[
p_{AL}^{\prime\prime}\left(x\right)=\frac{1}{2rs^{2}}\left(\frac{2\sinh\left(\frac{r}{s}\right)\sinh^{2}\left(\frac{x-m}{s}\right)}{\left(\cosh\left(\frac{x-m}{s}\right)+\cosh\left(\frac{r}{s}\right)\right)^{3}}-\frac{\sinh\left(\frac{r}{s}\right)\cosh\left(\frac{x-m}{s}\right)}{\left(\cosh\left(\frac{x-m}{s}\right)+\cosh\left(\frac{r}{s}\right)\right)^{2}}\right).
\]
Hence, 
\[
p_{AL}^{\prime}\left(a\right)=-p_{AL}^{\prime}\left(b\right)=\frac{1}{8rs}\tanh^{2}\left(\frac{r}{s}\right),
\]
\[
p_{AL}^{\prime\prime}\left(m\right)=-\frac{1}{2rs^{2}}\frac{\sinh\left(r/s\right)}{\left(1+\cosh\left(r/s\right)\right)^{2}}.
\]
Therefore, we have 
\begin{align*}
\left|p_{AL}^{\prime\prime}\left(m\right)\right|\left|\frac{a-b}{p_{AL}^{\prime}\left(a\right)-p_{AL}^{\prime}\left(b\right)}\right| & =\frac{4r}{s}\frac{\cosh^{2}\left(r/s\right)}{\sinh\left(r/s\right)\left(1+\cosh\left(r/s\right)\right)^{2}}\\
 & <\mathrm{4\left(r/s\right)csch}\left(r/s\right).
\end{align*}

% appendix D
\specialsection*{Appendix D. Condition for flat-topped $p_{BL}^{}$ }

The PDF $p_{BL}^{}\left(x\right)$ given by~(\ref{eq:pbl-def}) can
be rewritten as 
\[
p_{BL}^{}\left(x\mid a,b,s,t\right)=c\,F_{L}\!\left(\frac{x-a}{s}\right)F_{L}\!\left(\frac{b-x}{t}\right)
\]
where $F_{L}\left(x\right)$ is the abbreviation of $F_{L}\left(x;0,1\right)$. 
Since $F_{L}\left(-x\right)=1-F_{L}\left(x\right)$ and $F_{L}^{\prime}\left(x\right)=F_{L}\left(x\right)F_{L}\left(-x\right)$,
the first and second derivatives of $p_{BL}^{}\left(x\right)$ are expressed
as
\begin{align*}
p_{BL}^{\prime}\left(x\right) & =c\,F_{L}\!\left(\frac{x-a}{s}\right)F_{L}\!\left(\frac{b-x}{t}\right)\left[\frac{1}{s}F_{L}\!\left(\frac{a-x}{s}\right)-\frac{1}{t}F_{L}\!\left(\frac{x-b}{t}\right)\right]\\
 & =p_{BL}^{}\left(x\right)\left[\frac{1}{s}F_{L}\!\left(\frac{a-x}{s}\right)-\frac{1}{t}F_{L}\!\left(\frac{x-b}{t}\right)\right]
\end{align*}
\begin{align*}
p_{BL}^{\prime\prime}\left(x\right) & =p_{BL}^{\prime}\left(x\right)\left[\frac{1}{s}F_{L}\!\left(\frac{a-x}{s}\right)-\frac{1}{t}F_{L}\!\left(\frac{x-b}{t}\right)\right]\\
 & \quad-p_{BL}^{}\left(x\right)\left[\frac{1}{s^{2}}F_{L}\!\left(\frac{x-a}{s}\right)F_{L}\!\left(\frac{a-x}{s}\right)+\frac{1}{t^{2}}F_{L}\!\left(\frac{x-b}{t}\right)F_{L}\!\left(\frac{b-x}{t}\right)\right].
\end{align*}
Hence, we have
\[
p_{BL}^{\prime}\left(a\right)=\frac{c}{8}\left[\frac{1}{s}\left(1+\tanh\left(\frac{b-a}{2t}\right)\right)-\frac{1}{t}\left(1-\tanh^{2}\left(\frac{b-a}{2t}\right)\right)\right]
\]
\[
p_{BL}^{\prime}\left(b\right)=\frac{c}{8}\left[\frac{1}{s}\left(1-\tanh^{2}\left(\frac{b-a}{2s}\right)\right)-\frac{1}{t}\left(1+\tanh\left(\frac{b-a}{2s}\right)\right)\right]
\]
and the difference of $p_{BL}^{\prime}\left(a\right)-p_{BL}^{\prime}\left(b\right)$
satisfies the following inequality: 
\begin{align}
p_{BL}^{\prime}\left(a\right)-p_{BL}^{\prime}\left(b\right) & \geq\frac{c}{8}\left[\frac{1}{s}\tanh\left(\frac{b-a}{2t}\right)+\frac{1}{t}\tanh\left(\frac{b-a}{2s}\right)\right]\nonumber \\
 & \geq\frac{c}{4\sqrt{st}}\sqrt{\tanh\left(\frac{b-a}{2s}\right)\tanh\left(\frac{b-a}{2t}\right)}\nonumber \\
 & \geq\frac{c}{2\left[t\coth\left(\frac{b-a}{2s}\right)+s\coth\left(\frac{b-a}{2t}\right)\right]}\label{eq:ineq-pbl-d1-a-b}
\end{align}
where we have used the AM-GM inequality twice. As concerns
$p_{BL}^{\prime\prime}\left(x_{m}\right)$, it follows from $p_{BL}^{\prime}\left(x_{m}\right)=0$
that 
\[
\frac{1}{s}F_{L}\!\left(\frac{a-x_{m}}{s}\right)=\frac{1}{t}F_{L}\!\left(\frac{x_{m}-b}{t}\right)
\]
and hence 
\begin{align*}
p_{BL}^{\prime\prime}\left(x_{m}\right) & =-p_{BL}^{}\left(x_{m}\right)\left(\frac{F_{L}\!\left(\frac{x_{m}-a}{s}\right)F_{L}\!\left(\frac{x_{m}-b}{t}\right)+F_{L}\!\left(\frac{a-x_{m}}{s}\right)F_{L}\!\left(\frac{b-x_{m}}{t}\right)}{st}\right)\\
 & =-p_{BL}^{}\left(x_{m}\right)\frac{F_{L}\!\left(\frac{x_{m}-a}{s}\right)F_{L}\!\left(\frac{b-x_{m}}{t}\right)}{st}\left(\frac{F_{L}\!\left(\frac{x_{m}-b}{t}\right)}{F_{L}\!\left(\frac{b-x_{m}}{t}\right)}+\frac{F_{L}\!\left(\frac{a-x_{m}}{s}\right)}{F_{L}\!\left(\frac{x_{m}-a}{s}\right)}\right)\\
 & =-\frac{c\,F_{L}\!\left(\frac{x_{m}-a}{s}\right)^{2}F_{L}\!\left(\frac{b-x_{m}}{t}\right)^{2}}{st}\left(\exp\left(\frac{x_{m}-b}{t}\right)+\exp\left(\frac{a-x_{m}}{s}\right)\right).
\end{align*}
Let $y=\left(at+bs\right)/\left(s+t\right)$. Since $p_{BL}^{}\left(x\right)\leq p_{BL}^{}\left(x_{m}\right)$
holds for every $x$, we have 
\[
p_{BL}^{}\left(y\right)=\frac{c}{\left(1+\exp\left(\frac{a-b}{s+t}\right)\right)^{2}}\leq\frac{c}{\left(1+\exp\left(\frac{a-x_{m}}{s}\right)\right)\left(1+\exp\left(\frac{x_{m}-b}{t}\right)\right)}.
\]
It follows that 
\[
\exp\left(\frac{a-x_{m}}{s}\right)+\exp\left(\frac{x_{m}-b}{t}\right)<\left(2+\exp\left(\frac{a-b}{s+t}\right)\right)\exp\left(\frac{a-b}{s+t}\right)
\]
and therefore 
\[
\left|p_{BL}^{\prime\prime}\left(x_{m}\right)\right|<\frac{3c}{st}\exp\left(\frac{a-b}{s+t}\right).
\]
From this inequality and~(\ref{eq:ineq-pbl-d1-a-b}), we
have 
\[
\left|p_{BL}^{\prime\prime}\left(x_{m}\right)\right|\left|\frac{a-b}{p_{BL}^{\prime}\left(a\right)-p_{BL}^{\prime}\left(b\right)}\right|<\frac{6\left(\frac{b-a}{s}\coth\left(\frac{b-a}{2s}\right)+\frac{b-a}{t}\coth\left(\frac{b-a}{2t}\right)\right)}{\exp\left(\frac{b-a}{s+t}\right)}.
\]

% appendix E
\specialsection*{Appendix E. Approximation of the normalizing constant
of $p_{BL}^{}$ }

The normalization condition of $p_{B}^{}$ given by~(\ref{eq:pb-def})
can be expressed as 
\begin{align*}
1/c & =\int_{-\infty}^{\infty}F\left(x;a,s\right)\left(1-G\left(x;b,t\right)\right)dx\\
 & =\int_{-\infty}^{\infty}\left\{ F\left(x;a,s\right)-F\left(x;b,s\right)+\left(1-F\left(x;a,s\right)\right)G\left(x;b,t\right)\right\} dx\\
 & =b-a+\int_{-\infty}^{\infty}\left(1-F\left(x;a,s\right)\right)G\left(x;b,t\right)dx
\end{align*}
if $\int_{-\infty}^{\infty}\left(F\left(x;b,s\right)-G\left(x;b,t\right)\right)dx=0$.
Let $\delta$ be the last term of the integral such that 
\begin{align*}
\delta & =\int_{-\infty}^{\infty}\left(1-F\left(x;a,s\right)\right)G\left(x;b,t\right)dx\\
 & <\min_{y}\left\{ \int_{-\infty}^{y}G\left(x;b,t\right)dx+\int_{y}^{\infty}\left(1-F\left(x;a,s\right)\right)dx\right\} .
\end{align*}
For $p_{BL}^{}\left(x\right)=F_{L}\left(x;a,s\right)\left(1-F_{L}\left(x;b,t\right)\right)$,
letting $y$ be a point such that $F_{L}\left(y;a,s\right)=1-F_{L}\left(y;b,t\right)$
gives 
\[
\delta<t\ln\left(1+\exp\left(\frac{y-b}{t}\right)\right)+s\ln\left(1+\exp\left(\frac{a-y}{s}\right)\right),
\]
If $p_{BL}^{}$ is \textit{flat-topped} under the condition $\exp\left(\left(a-y\right)/s\right),\exp\left(\left(y-b\right)/t\right)<\varepsilon\ll1$,
that is $1-F_{L}\left(y;a,s\right)=F_{L}\left(y;b,t\right)<\varepsilon$,
then it can be approximated by (\ref{eq:pbl-flat-approx}) and $\delta$
must be very small so that the error of the approximation can be less
influential.

% appendix F
\specialsection*{Appendix F. Integration of $p_{CF}^{}$ and $p_{CH}^{}$ }

Let $p_{CF}^{}$ be a PDF defined by 
\[
p_{CF}^{}\left(x\right)=c_{F}\left\{ 1+\exp\left(\frac{\left|x\right|^{\beta}-r^{\beta}}{s^{\beta}}\right)\right\} ^{-1}
\]
where $c_{F},r,s,\beta>0$ are constants. Let $k$ be a non-negative even integer. The $k$-th central moment of $p_{CF}^{}$ is given by 
\begin{align*}
\int_{-\infty}^{\infty}x^{k}p_{CF}^{}\left(x\right)dx & =2c_{F}\int_{0}^{\infty}x^{k}\left\{ 1+\exp\left(\frac{x^{\beta}-r^{\beta}}{s^{\beta}}\right)\right\} ^{-1}dx\\
 & =c_{F}\frac{2s}{\beta}\int_{0}^{\infty}\frac{s^{k}u^{k/\beta}u^{1/\beta-1}}{1+\exp\left(u-r^{\beta}/s^{\beta}\right)}\mathrm{d}u\\
 & =c_{F}\frac{2s^{k+1}}{\beta}\Gamma\left(\frac{k+1}{\beta}\right)F_{\frac{k+1}{\beta}-1}\left(\frac{r^{\beta}}{s^{\beta}}\right).
\end{align*}
where $u=x^{\beta}/s^{\beta}$, $\Gamma$ is the gamma function, and
$F_{j}\left(\cdot\right)$ is the complete Fermi-Dirac integral given
by~(\ref{eq:FD-integ}). It follows from the normalization condition
for $k=0$ and $\Gamma\left(x+1\right)=x\Gamma\left(x\right)$ that
\[
c_{F}=\left\{ 2s\Gamma\left(\frac{1}{\beta}+1\right)F_{\frac{1}{\beta}-1}\left(\frac{r^{\beta}}{s^{\beta}}\right)\right\} ^{-1}.
\]
The CDF of $p_{CF}^{}$ is expressed as 
\[
P_{CF}\left(x\right)=\frac{1}{2}\left(1+\mathrm{sgn}\left(x\right)\left\{ 1-\frac{F_{1/\beta-1}\left(r^{\beta}/s^{\beta},\left|x\right|^{\beta}/s^{\beta}\right)}{F_{1/\beta-1}\left(r^{\beta}/s^{\beta}\right)}\right\} \right)
\]
where 
\[
F_{j}\left(x,u\right)=\frac{1}{\Gamma\left(j+1\right)}\int_{u}^{\infty}\frac{t^{j}}{e^{t-x}+1}dt
\]
is the incomplete Fermi\textendash Dirac integral for an index $j$.
If $\beta=1$, then $F_{0}\left(x,u\right)=\ln\left(1+\exp\left(x-u\right)\right)$
and we have 
\[
P_{CF}\left(x\right)=\frac{1}{2}\left(1+\mathrm{sgn}\left(x\right)\left\{ 1-\frac{\ln\left(1+\exp\left(\frac{r-\left|x\right|}{s}\right)\right)}{\ln\left(1+\exp\left(\frac{r}{s}\right)\right)}\right\} \right).
\]

Let $p_{CH}^{}$ be a PDF defined by 
\[
p_{CH}^{}\left(x\right)=\frac{c_{H}\,\sinh\left(r^{\beta}/s^{\beta}\right)}{\cosh\left(\left|x\right|^{\beta}/s^{\beta}\right)+\cosh\left(r^{\beta}/s^{\beta}\right)}.
\]
As in $p_{CF}^{}$, the $k$-th central moment of $p_{CH}^{}$ is given by 
\begin{align*}
\int_{-\infty}^{\infty}x^{k}p_{CH}^{}\left(x\right)dx & =2\,c_{H}\int_{0}^{\infty}\left\{ \frac{x^{k}}{1+\exp\left(\frac{x^{\beta}-r^{\beta}}{s^{\beta}}\right)}-\frac{x^{k}}{1+\exp\left(\frac{x^{\beta}+r^{\beta}}{s^{\beta}}\right)}\right\} dx\\
 & =c_{H}\frac{2s^{k+1}}{\beta}\Gamma\left(\frac{k+1}{\beta}\right)\left\{ F_{\frac{k+1}{\beta}-1}\left(\frac{r^{\beta}}{s^{\beta}}\right)-F_{\frac{k+1}{\beta}-1}\left(-\frac{r^{\beta}}{s^{\beta}}\right)\right\} .
\end{align*}
where 
\[
c_{H}=\left[2\,s\,\Gamma\left(\frac{1}{\beta}+1\right)\left\{ F_{\frac{1}{\beta}-1}\left(\frac{r^{\beta}}{s^{\beta}}\right)-F_{\frac{1}{\beta}-1}\left(-\frac{r^{\beta}}{s^{\beta}}\right)\right\} \right]^{-1}.
\]
The CDF of $p_{CH}^{}$ is expressed as 
\[
P_{CH}\left(x\right)=\frac{1}{2}\left(1+\mathrm{sgn}\left(x\right)\left\{ 1-\frac{F_{\frac{1}{\beta}-1}\left(\frac{r^{\beta}}{s^{\beta}},\frac{\left|x\right|^{\beta}}{s^{\beta}}\right)-F_{\frac{1}{\beta}-1}\left(-\frac{r^{\beta}}{s^{\beta}},\frac{\left|x\right|^{\beta}}{s^{\beta}}\right)}{F_{\frac{1}{\beta}-1}\left(\frac{r^{\beta}}{s^{\beta}}\right)-F_{\frac{1}{\beta}-1}\left(-\frac{r^{\beta}}{s^{\beta}}\right)}\right\} \right).
\]
If $\beta=1$, then we have
\[
P_{CH}\left(x\right)=\frac{s}{2r}\ln\left(\frac{1+\exp\left(\frac{x+r}{s}\right)}{1+\exp\left(\frac{x-r}{s}\right)}\right).
\]

% appendix G
\specialsection*{Appendix G. Integration of $p_{CM}^{}$ and $p_{CL}^{}$}

Let $p_{C}^{}$ be a PDF for $n$-multivariate distribution defined by
\[
p_{C}^{}\left(\mathbf{x}\mid\mathbf{m},\mathbf{\boldsymbol{\Sigma}}\right)=\frac{c}{h+g\left(\left\{ \left(\mathbf{x}-\mathbf{m}\right)^{\mathsf{T}}\mathbf{\boldsymbol{\Sigma}}^{-1}\left(\mathbf{x}-\mathbf{m}\right)\right\} ^{n/2}\right)}
\]
where $\mathbf{x}$ and $\mathbf{m}$ are $n$ dimensional vectors
and $\boldsymbol{\Sigma}$ is an $n\times n$ positive-definite matrix.
Based on the eigendecomposition of $\boldsymbol{\Sigma}$ with an
orthogonal matrix $\mathbf{Q}$ such that $\boldsymbol{\Sigma}^{-1}=\mathbf{Q}\mathbf{\boldsymbol{\Lambda}}^{-1}\mathbf{Q}^{-1}$
where $\mathbf{\boldsymbol{\Lambda}}$ is a diagonal matrix and $\left|\boldsymbol{\Sigma}\right|=\left|\mathbf{\boldsymbol{\Lambda}}\right|=\prod_{i=1}^{n}\lambda_{i}$,
we make the changes of variables $\mathbf{y}=\mathbf{Q}^{-1}\left(\mathbf{x}-\mathbf{m}\right)$
and $z_{i}=y_{i}/\sqrt{\lambda_{i}}$ that satisfy 
\[
\left(\mathbf{x}-\mathbf{m}\right)^{\mathsf{T}}\mathbf{\boldsymbol{\Sigma}}^{-1}\left(\mathbf{x}-\mathbf{m}\right)=\mathbf{y}^{\mathsf{T}}\mathbf{\boldsymbol{\Lambda}}^{-1}\mathbf{y}=\sum_{i=1}^{n}z_{i}^{2}.
\]
The Jacobian determinant of the transformation from $\mathbf{x}$
to $\mathbf{y}$ is 1 and that from $\mathbf{y}$ to $\mathbf{z}$
is $\prod_{i=1}^{n}\sqrt{\lambda_{i}}=\sqrt{\left|\boldsymbol{\Sigma}\right|}$.
The integral of $p_{C}^{}$ for $n\geq2$ can be evaluated by
using the further change of variables from Cartesian to polar coordinates
as follows: 
\begin{align}
 & \int_{\mathbb{R}^{n}}p_{C}^{}\left(\mathbf{x}\mid\mathbf{m},\boldsymbol{\Sigma}\right)d\mathbf{x}\nonumber \\
 & =\intop_{-\infty}^{\infty}\cdots\intop_{-\infty}^{\infty}\frac{c}{h+g\left(\left(\sum_{i=1}^{n}z_{i}^{2}\right)^{n/2}\right)}\sqrt{\lambda_{1}\cdots\lambda_{n}}\,dz_{1}\cdots dz_{n}\nonumber \\
 & =\left|\boldsymbol{\Sigma}\right|^{1/2}\intop_{0}^{2\pi}\left(\intop_{0}^{\pi}\cdots\intop_{0}^{\pi}\left(\intop_{0}^{\infty}\frac{c}{h+g\left(\rho^{n}\right)}J_{n}\,d\rho\right)d\varphi_{1}\cdots d\varphi_{n-2}\right)d\varphi_{n-1}\nonumber \\
 & =\left|\boldsymbol{\Sigma}\right|^{1/2}\left(2\pi\prod_{i=1}^{n-2}\int_{0}^{\pi}\sin^{n-1-i}\varphi_{i}\,d\varphi_{i}\right)\left(\int_{0}^{\infty}\frac{c\rho^{n-1}}{h+g\left(\rho^{n}\right)}d\rho\right)\nonumber \\
 & =\left|\boldsymbol{\Sigma}\right|^{1/2}\left(\frac{2\pi^{n/2}}{\Gamma\left(n/2\right)}\right)\left(\frac{1}{n}\int_{0}^{\infty}\frac{c}{h+g\left(u\right)}du\right)\nonumber \\
 & =\frac{\pi^{n/2}\left|\boldsymbol{\Sigma}\right|^{1/2}}{\Gamma\left(n/2+1\right)}\int_{0}^{\infty}\frac{c}{h+g\left(u\right)}du\label{eq:pc-g-func} 
\end{align}
where $z_{1}=\rho\cos\left(\varphi_{1}\right)$, $z_{2}=\rho\sin\left(\varphi_{1}\right)\cos\left(\varphi_{2}\right)$,
$z_{3}=\rho\sin\left(\varphi_{1}\right)\sin\left(\varphi_{2}\right)\cos\left(\varphi_{3}\right)$,
$\ldots$, $z_{n-1}=\rho\sin\left(\varphi_{1}\right)\cdots\sin\left(\varphi_{n-2}\right)\cos\left(\varphi_{n-1}\right)$,
$z_{n}=\rho\sin\left(\varphi_{1}\right)\cdots\sin\left(\varphi_{n-2}\right)\sin\left(\varphi_{n-1}\right)$,
$u=\rho^{n}$, $J_{n}$ is the Jacobian determinant such that $J_{n}=\rho^{n-1}\prod_{i=1}^{n-2}\sin^{n-1-i}\varphi_{i}$
and the integral with respect to $\varphi_{1},\ldots,\varphi_{n-1}$
is represented by the following special functions: 
\[
\prod_{i=1}^{n-2}\int_{0}^{\pi}\sin^{n-1-i}\varphi_{i}d\varphi_{i}=\prod_{i=1}^{n-2}\mathrm{B}\left(\frac{n-i}{2},\frac{1}{2}\right)=\prod_{i=1}^{n-2}\frac{\Gamma\left(\frac{n-i}{2}\right)\Gamma\left(\frac{1}{2}\right)}{\Gamma\left(\frac{n+1-i}{2}\right)}=\frac{\pi^{n/2-1}}{\Gamma\left(n/2\right)}.
\]

The integral of $p_{CM}^{}$ can be obtained by substituting $g\left(u\right)=\exp\left(ut-r^{n}t\right)$
and $h=1$ into~(\ref{eq:pc-g-func}) as follows: 
\[
\int_{0}^{\infty}\frac{c}{h+g\left(u\right)}du=\int_{0}^{\infty}\frac{c}{1+\exp\left(ut-r^{n}t\right)}du=\frac{c}{t}F_{0}\left(r^{n}t\right)=\frac{c}{t}\ln\left(1+\exp\left(r^{n}t\right)\right).
\]
In much the same way, the integral of $p_{CL}^{}$ can be obtained by
substituting $g\left(u\right)=\cosh\left(ut\right)/\sinh\left(r^{n}t\right)$
and $h=\coth\left(r^{n}t\right)$ into~(\ref{eq:pc-g-func}) as follows:
\begin{align*}
\int_{0}^{\infty}\frac{c}{h+g\left(u\right)}du & =\int_{0}^{\infty}\frac{c\sinh\left(r^{n}t\right)}{\cosh\left(r^{n}t\right)+\cosh\left(ut\right)}du\\
 & =\int_{0}^{\infty}\left(\frac{c}{1+\exp\left(ut-r^{n}t\right)}-\frac{c}{1+\exp\left(ut+r^{n}t\right)}\right)du\\
 & =\frac{c}{t}\left\{ F_{0}\left(r^{n}t\right)-F_{0}\left(-r^{n}t\right)\right\} \\
 & =\frac{c}{t}\left\{ \ln\left(1+\exp\left(r^{n}t\right)\right)-\ln\left(1+\exp\left(-r^{n}t\right)\right)\right\} \\
 & =c\,r^{n}.
\end{align*}

% appendix H
\specialsection*{Appendix H. Partial derivatives of $l_{AL}$ }

The PDF $p_{AL}^{}$ given by~(\ref{eq:pal-def}) is rewritten as 
\begin{align*}
p_{AL}^{}\left(x\mid a,b,s\right) & =\frac{1}{2\left(b-a\right)}\left(\tanh\left(\frac{x-a}{2s}\right)-\tanh\left(\frac{x-b}{2s}\right)\right)\\
 & =\frac{1}{2\left(b-a\right)}\left(\frac{\sinh\left(\frac{b-a}{2s}\right)}{\cosh\left(\frac{x-a}{2s}\right)\cosh\left(\frac{x-b}{2s}\right)}\right).
\end{align*}
The log-likelihood $l_{AL}$ for an i.i.d. sample $\left\{ x_{1},\ldots,x_{N}\right\} $
is 
\begin{align*}
l_{AL} & =\sum_{i=1}^{N}\ln p_{AL}^{}\left(x_{i}\mid a,b,s\right)\\
 & =N\ln\left(\frac{\sinh\left(\frac{b-a}{2s}\right)}{2\left(b-a\right)}\right)-\sum_{i=1}^{N}\ln\left\{ \cosh\left(\frac{x_{i}-a}{2s}\right)\cosh\left(\frac{x_{i}-b}{2s}\right)\right\} .
\end{align*}
Thus the partial derivatives of the log-likelihood are as follows:
\[
\partial_{a}l_{AL}=\frac{N}{b-a}-\frac{N}{2s}\coth\left(\frac{b-a}{2s}\right)+\frac{1}{2s}\sum_{i=1}^{N}\tanh\left(\frac{x_{i}-a}{2s}\right),
\]
\[
\partial_{b}l_{AL}=-\frac{N}{b-a}+\frac{N}{2s}\coth\left(\frac{b-a}{2s}\right)+\frac{1}{2s}\sum_{i=1}^{N}\tanh\left(\frac{x_{i}-b}{2s}\right),
\]
\begin{align*}
\partial_{s}l_{AL}= & -\frac{N}{s}\left(\frac{b-a}{2s}\right)\coth\left(\frac{b-a}{2s}\right)\\
 & \hspace{5bp}+\frac{1}{s}\sum_{i=1}^{N}\left\{ \left(\frac{x_{i}-a}{2s}\right)\tanh\left(\frac{x_{i}-a}{2s}\right)+\left(\frac{x_{i}-b}{2s}\right)\tanh\left(\frac{x_{i}-b}{2s}\right)\right\} ,
\end{align*}
\[
\partial_{aa}l_{AL}=\frac{N}{\left(b-a\right)^{2}}-\frac{N}{4s^{2}}\mathrm{csch}^{2}\left(\frac{b-a}{2s}\right)-\frac{1}{4s^{2}}\sum_{i=1}^{N}\mathrm{sech}^{2}\left(\frac{x_{i}-a}{2s}\right),
\]
\[
\partial_{bb}l_{AL}=\frac{N}{\left(b-a\right)^{2}}-\frac{N}{4s^{2}}\mathrm{csch}^{2}\left(\frac{b-a}{2s}\right)-\frac{1}{4s^{2}}\sum_{i=1}^{N}\mathrm{sech}^{2}\left(\frac{x_{i}-b}{2s}\right),
\]
\begin{align*}
\partial_{ss}l_{AL}= & \frac{N}{s^{2}}\left\{ \left(\frac{b-a}{2s}\right)\coth\left(\frac{b-a}{2s}\right)-\left(\frac{b-a}{2s}\right)^{2}\mathrm{csch}^{2}\left(\frac{b-a}{2s}\right)\right\} \\
 & -\frac{1}{s^{2}}\sum_{i=1}^{N}\left\{ \left(\frac{x_{i}-a}{2s}\right)\tanh\left(\frac{x_{i}-a}{2s}\right)+\left(\frac{x_{i}-a}{2s}\right)^{2}\mathrm{sech}^{2}\left(\frac{x_{i}-a}{2s}\right)\right\} \\
 & -\frac{1}{s^{2}}\sum_{i=1}^{N}\left\{ \left(\frac{x_{i}-b}{2s}\right)\tanh\left(\frac{x_{i}-b}{2s}\right)+\left(\frac{x_{i}-b}{2s}\right)^{2}\mathrm{sech}^{2}\left(\frac{x_{i}-b}{2s}\right)\right\} ,
\end{align*}
\[
\partial_{ab}l_{AL}=-\frac{N}{\left(b-a\right)^{2}}+\frac{N}{4s^{2}}\mathrm{csch}^{2}\left(\frac{b-a}{2s}\right),\hspace{3.8cm}
\]
\begin{align*}
\partial_{as}l_{AL}= & \frac{N}{2s^{2}}\left\{ \coth\left(\frac{b-a}{2s}\right)-\left(\frac{b-a}{2s}\right)\mathrm{csch}^{2}\left(\frac{b-a}{2s}\right)\right\} \\
 & -\frac{1}{2s^{2}}\sum_{i=1}^{N}\left\{ \tanh\left(\frac{x_{i}-a}{2s}\right)+\left(\frac{x_{i}-a}{2s}\right)\mathrm{sech}^{2}\left(\frac{x_{i}-a}{2s}\right)\right\} ,
\end{align*}
\begin{align*}
\partial_{bs}l_{AL}= & -\frac{N}{2s^{2}}\left\{ \coth\left(\frac{b-a}{2s}\right)-\left(\frac{b-a}{2s}\right)\mathrm{\mathrm{csch}^{2}}\left(\frac{b-a}{2s}\right)\right\} \\
 & -\frac{1}{2s^{2}}\sum_{i=1}^{N}\left\{ \tanh\left(\frac{x_{i}-b}{2s}\right)+\left(\frac{x_{i}-b}{2s}\right)\mathrm{sech}^{2}\left(\frac{x_{i}-b}{2s}\right)\right\} .
\end{align*}

% appendix I
\specialsection*{Appendix I. KL divergence between $p_{MU}^{}$ and $p_{MN}^{}$ }

The variance $\sigma_{11}$ of $p_{MU}^{}\left(\mathbf{x}\mid\mathbf{0},r\right)$
given by~(\ref{eq:pmu-def}) is evaluated in much the same way as in~(\ref{eq:pc-g-func})
:
\begin{align*}
\sigma_{11} & =\int_{-\infty}^{\infty}\cdots\int_{-\infty}^{\infty}x_{1}^{2}\:p_{MU}^{}\left(\mathbf{x}\mid\mathbf{0},r\right)dx_{1}\cdots dx_{n}\\
 & =\intop_{0}^{2\pi}\intop_{0}^{\pi}\cdots\intop_{0}^{\pi}\intop_{0}^{r}\rho^{2}\cos^{2}\varphi_{1}\:q_{MU}^{}\left(\rho\right)J_{n}\,d\rho\,d\varphi_{1}\cdots d\varphi_{n-1}\\
 & =\frac{\Gamma\left(\frac{n}{2}+1\right)}{\pi^{n/2}r^{n}}2\pi\mathrm{B}\left(\frac{n-1}{2},\frac{3}{2}\right)\left(\prod_{i=2}^{n-2}\mathrm{B}\left(\frac{n-i}{2},\frac{1}{2}\right)\right)\intop_{0}^{r}\rho^{n+1}d\rho\\
 & =\frac{\Gamma\left(\frac{n}{2}+1\right)}{\pi^{n/2}r^{n}}2\pi\frac{\Gamma\left(\frac{n-1}{2}\right)\Gamma\left(\frac{3}{2}\right)}{\Gamma\left(\frac{n+2}{2}\right)}\left(\frac{\Gamma\left(\frac{1}{2}\right)^{n-3}}{\Gamma\left(\frac{n-1}{2}\right)}\right)\frac{r^{n+2}}{n+2}\\
 & =\frac{r^{2}}{n+2}
\end{align*}
where $p_{MU}^{}\left(\mathbf{x}\mid\mathbf{0},r\right)$ is transformed
into the radial function 
\[
q_{MU}^{}\left(\rho\right)=\begin{cases}
\frac{\Gamma\left(n/2+1\right)}{\pi^{n/2}r^{n}} & \rho\leq r,\\
0 & \mathrm{otherwise}.
\end{cases}
\]
Hence, the fitted normal distribution $p_{MN}^{\ast}\left(\mathbf{x}\mid\mathbf{0},\hat{\boldsymbol{\Sigma}}\right)$
is expressed by the following radial function with respect to $\rho=\left(\mathbf{x}^{\mathsf{T}}\mathbf{x}\right)^{1/2}$:
\[
q_{MN}^{}\left(\rho\right)=\left(\frac{n+2}{2\pi r^{2}}\right)^{n/2}\exp\left(-\frac{n+2}{2r^{2}}\rho^{2}\right).
\]
The expected log-likelihoods of $p_{MU}^{}$ and $p_{MN}^{\ast}$ are 
\begin{align*}
 & \int_{-\infty}^{\infty}\cdots\int_{-\infty}^{\infty}p_{MU}^{}\left(\mathbf{x}\mid\mathbf{0},r\right)\ln p_{MU}^{}\left(\mathbf{x}\mid\mathbf{0},r\right)dx_{1}\cdots dx_{n}\\
 & =\intop_{0}^{2\pi}\intop_{0}^{\pi}\cdots\intop_{0}^{\pi}\intop_{0}^{\infty}q_{MU}^{}\left(\rho\right)\ln q_{MU}^{}\left(\rho\right)J_{n}\,d\rho\,d\varphi_{1}\cdots d\varphi_{n-1}\\
 & =\ln\frac{\Gamma\left(n/2+1\right)}{\pi^{n/2}r^{n}},
\end{align*}
\begin{align*}
 & \int_{-\infty}^{\infty}\cdots\int_{-\infty}^{\infty}p_{MU}^{}\left(\mathbf{x}\mid\mathbf{0},r\right)\ln p_{MN}^{\ast}\left(\mathbf{x}\mid\mathbf{0},\hat{\boldsymbol{\Sigma}}\right)dx_{1}\cdots dx_{n}\\
 & =\intop_{0}^{2\pi}\intop_{0}^{\pi}\cdots\intop_{0}^{\pi}\intop_{0}^{\infty}q_{MU}^{}\left(\rho\right)\ln q_{MN}^{}\left(\rho\right)J_{n}\,d\rho\,d\varphi_{1}\cdots d\varphi_{n-1}\\
 & =\intop_{0}^{2\pi}\intop_{0}^{\pi}\cdots\intop_{0}^{\pi}\intop_{0}^{r}\frac{\Gamma\left(n/2+1\right)}{\pi^{n/2}r^{n}}\left(\frac{n}{2}\ln\frac{n+2}{2\pi r^{2}}-\frac{n+2}{2r^{2}}\rho^{2}\right)J_{n}\,d\rho\,d\varphi_{1}\cdots d\varphi_{n-1}\\
 & =\frac{n}{2}\ln\frac{n+2}{2\pi r^{2}}-\frac{n}{2}.
\end{align*}
Therefore the KL divergence is 
\begin{align*}
D_{KL}\left(p_{MU}^{}\parallel p_{MN}^{\ast}\right) & =\ln\frac{\Gamma\left(n/2+1\right)}{\pi^{n/2}r^{n}}-\left(\frac{n}{2}\ln\frac{n+2}{2\pi r^{2}}-\frac{n}{2}\right)\\
 & =\ln\Gamma\left(\frac{n}{2}+1\right)-\frac{n}{2}\ln\left(\frac{n}{2}+1\right)+\frac{n}{2}
\end{align*}
and the $L_{1}$ distance is
\begin{align*}
 & D_{L_{1}}\left(p_{MU}^{},p_{MN}^{\ast}\right)\\
 & =\int_{-\infty}^{\infty}\cdots\int_{-\infty}^{\infty}\left|p_{MU}^{}\left(\mathbf{x}\mid\mathbf{0},r\right)-p_{MN}^{\ast}\left(\mathbf{x}\mid\mathbf{0},\hat{\boldsymbol{\Sigma}}\right)\right|dx_{1}\cdots dx_{n}\\
 & =\int_{0}^{2\pi}\int_{0}^{\pi}\cdots\int_{0}^{\pi}\int_{0}^{\infty}\left|q_{MU}^{}\left(\rho\right)-q_{MN}^{}\left(\rho\right)\right|J_{n}\,d\rho\,d\varphi_{1}\cdots d\varphi_{n-1}\\
 & =I_{\varphi}\left\{ \int_{0}^{u}\left(q_{MN}^{}-q_{MU}^{}\right)\rho^{n-1}d\rho+\int_{u}^{r}\left(q_{MU}^{}-q_{MN}^{}\right)\rho^{n-1}d\rho+\int_{r}^{\infty}q_{MN}^{}\rho^{n-1}d\rho\right\} \\
 & =\left(1-2\frac{u^{n}}{r^{n}}\right)+\frac{1}{\Gamma\left(n/2\right)}\left\{ \Gamma\left(n/2\right)-2\Gamma\left(\frac{n}{2},\frac{n+2}{2r^{2}}u^{2}\right)+2\Gamma\left(\frac{n}{2},\frac{n+2}{2r^{2}}r^{2}\right)\right\} \\
 & =2\left\{ 1-\chi_{n}^{n}-\frac{\Gamma\left(n/2,\left(n/2+1\right)\chi_{n}^{2}\right)-\Gamma\left(n/2,n/2+1\right)}{\Gamma\left(n/2\right)}\right\} .
\end{align*}
where $\Gamma\left(s,x\right)=\int_{x}^{\infty}t^{s-1}e^{-t}dx$ is
the upper incomplete gamma function and $u=\chi_{n}r$ satisfies the
equation $q_{MU}^{}\left(u\right)=q_{MN}^{}\left(u\right)$, i.e.,
\[
\chi_{n}=\frac{u}{r}=\sqrt{\frac{1}{\left(n/2+1\right)}\ln\frac{\left(n/2+1\right)^{n/2}}{\Gamma\left(n/2+1\right)}}.
\]
The partial integrals are evaluated as follows:
\[
I_{\varphi}=\intop_{0}^{2\pi}\left(\intop_{0}^{\pi}\cdots\intop_{0}^{\pi}\left(\prod_{i=1}^{n-2}\sin^{n-1-i}\varphi_{i}\right)d\varphi_{1}\cdots d\varphi_{n-2}\right)d\varphi_{n-1}=\frac{2\pi^{n/2}}{\Gamma\left(n/2\right)},
\]
\[
\int_{u}^{r}q_{MU}^{}\left(\rho\right)\rho^{n-1}d\rho=\int_{u}^{r}\frac{\Gamma\left(n/2+1\right)}{\pi^{n/2}r^{n}}\rho^{n-1}d\rho=\frac{\Gamma\left(n/2\right)}{2\pi^{n/2}r^{n}}\left(r^{n}-u^{n}\right),
\]
\begin{align*}
\int_{u}^{r}q_{MN}\left(\rho\right)\rho^{n-1}d\rho & =\int_{u}^{r}\left(\frac{n+2}{2\pi r^{2}}\right)^{n/2}\exp\left(-\frac{n+2}{2r^{2}}\rho^{2}\right)\rho^{n-1}d\rho\\
 & =\frac{1}{2\pi^{n/2}}\int_{\left(n/2+1\right)u^{2}/r^{2}}^{n/2+1}t^{n/2-1}e^{-t}\,dt\\
 & =\frac{1}{2\pi^{n/2}}\left\{ \Gamma\left(\frac{n}{2},\left(\frac{n}{2}+1\right)\frac{u^{2}}{r^{2}}\right)-\Gamma\left(\frac{n}{2},\frac{n}{2}+1\right)\right\} .
\end{align*}
The $L_{1}$ distance $D_{L_{1}}\left(p_{MU}^{},p_{MN}^{\ast}\right)$ increases
with $n$, depending on the characteristics of $\chi_{n}$ such that
$0<\chi_{n}<1$, $\lim_{n\rightarrow\infty}\chi_{n}^{2}=1$, and $\lim_{n\rightarrow\infty}\chi_{n}^{n}=0$.

%\specialsection*{Acknowledgment}
\specialsection*{Acknowledgment}

This work was supported by JSPS KAKENHI Grant Number JP18K11603. Figures
have been produced using PGFPLOTS~\cite{Feuersanger} and matplotlib~\cite{Hunter}.

%\begin{thebibliography}{10}

\end{document}